%% file: TOSEM.tex
\documentclass[acmsmall]{acmart}

\AtBeginDocument{%
  \providecommand\BibTeX{{%
    Bib\TeX}}}

\setcopyright{acmlicensed}
\acmJournal{TOSEM}
\acmYear{2025} \acmVolume{1} \acmNumber{1} \acmArticle{1} \acmMonth{1}
\acmDOI{10.1145/3731556}

\usepackage{booktabs} 
\usepackage{threeparttable}
\usepackage{caption} 
\usepackage{tabularx}

\usepackage{amsmath,amsfonts}
\usepackage{algorithmic}
\usepackage{graphicx}
\usepackage{textcomp}
\usepackage{xcolor}
\usepackage{tcolorbox}
\usepackage{hyperref}
\usepackage{multirow}
\usepackage{amsthm}
\newtheorem{definition}{Definition}
\usepackage{pifont}
\usepackage{xspace}
\usepackage{caption}
\newcommand{\tool}{NeuSemSlice\xspace}     
\def\BibTeX{{\rm B\kern-.05em{\sc i\kern-.025em b}\kern-.08em
    T\kern-.1667em\lower.7ex\hbox{E}\kern-.125emX}}
\makeatletter
\DeclareRobustCommand\onedot{\futurelet\@let@token\@onedot}
\def\@onedot{\ifx\@let@token.\else.\null\fi\xspace}
\def\eg{\emph{e.g}\onedot} 
\def\ie{\emph{i.e}\onedot}

\makeatother

\newcommand{\zsd}[1] {{\color{black}{#1}}}

\usepackage[linesnumbered,ruled,vlined]{algorithm2e}
\usepackage{enumitem}

\definecolor{darkpurple}{RGB}{126,100,158}
\definecolor{darkgreen}{RGB}{120,148,64}
\definecolor{lightblue}{RGB}{146,205,220}
\ccsdesc[500]{Software and its engineering~Maintaining software}
\ccsdesc[300]{Computing methodologies~Artificial intelligence}

\begin{document}

\title{\tool{}: Towards Effective DNN Model Maintenance via Neuron-level Semantic Slicing}

\author{Shide Zhou}
\affiliation{%
  \institution{Huazhong University of Science and Technology}
  \city{Wuhan}
  \country{China}}
\email{shidez@hust.edu.cn}

\author{Tianlin Li}
\authornote{Corresponding author.}
\affiliation{%
  \institution{Nanyang Technological University}
  \city{Singapore}
  \country{Singapore}}
\email{tianlin001@e.ntu.edu.sg}

\author{Yihao Huang}
\affiliation{%
  \institution{Nanyang Technological University}
  \city{Singapore}
  \country{Singapore}}
\email{huang.yihao@ntu.edu.sg}

\author{Ling Shi}
\affiliation{%
  \institution{Nanyang Technological University}
  \city{Singapore}
  \country{Singapore}}
\email{ling.shi@ntu.edu.sg}

\author{Kailong Wang}
\authornotemark[1]
\affiliation{%
  \institution{Huazhong University of Science and Technology}
  \city{Wuhan}
  \country{China}}
\email{wangkl@hust.edu.cn}

\author{Yang Liu}
\affiliation{%
  \institution{Nanyang Technological University}
  \city{Singapore}
  \country{Singapore}}
\email{yangliu@ntu.edu.sg}

\author{Haoyu Wang}
\affiliation{%
  \institution{Huazhong University of Science and Technology}
  \city{Wuhan}
  \country{China}}
\email{haoyuwang@hust.edu.cn}

\keywords{Deep Neural Networks, Model Maintenance}

\begin{abstract}
Deep Neural networks~(DNNs), extensively applied across diverse disciplines, are characterized by their integrated and monolithic architectures, setting them apart from conventional software systems.
This architectural difference introduces particular challenges to maintenance tasks, such as model restructure~(\eg, model compression), re-adaptation~(\eg, fitting new samples), and incremental development~(\eg, continual knowledge accumulation). 

Prior research addresses these challenges by identifying task-critical neuron layers, and dividing neural networks into semantically-similar sequential modules. 
However, such layer-level approaches fail to precisely identify and manipulate neuron-level semantic components, restricting their applicability to finer-grained model maintenance tasks. 

In this work, we implement \tool{}, a novel framework that introduces the \textit{semantic slicing} technique to effectively identify critical neuron-level semantic components in DNN models for \textit{semantic-aware model maintenance} tasks. Specifically, \textit{semantic slicing} identifies, categorizes and merges critical neurons across different categories and layers according to their semantic similarity, enabling their flexibility and effectiveness in the subsequent tasks.

For \textit{semantic-aware model maintenance} tasks, we provide a series of novel strategies based on \textit{semantic slicing} to enhance \tool{}. They include semantic components~(\ie, critical neurons) preservation for model restructure, critical neuron tuning for model re-adaptation, and non-critical neuron training for model incremental development. A thorough evaluation has demonstrated that \tool{} significantly outperforms baselines in all three tasks.
\end{abstract}

\maketitle
\input{Chapters/intro}
\input{Chapters/background}
\input{Chapters/motivation}
\input{Chapters/methodology}
\input{Chapters/evaluation}
\input{Chapters/conclusion}
\bibliographystyle{ACM-Reference-Format}
\bibliography{paper}

\end{document}

%% file: Chapters/intro.tex
\section{Introduction}

In the evolving landscape of artificial intelligence, DNNs represent a dynamic shift towards data-driven programming paradigms, extensively adopted across various fields~\cite{kollias2018deep,bansal2018chauffeurnet,theate2021application}. As these models increasingly serve as core components in modern software, their maintenance transcends mere optimization challenges and requires strategies grounded in established software engineering principles~\cite{10.1145/3604609, 10.1145/3688841, ren2023deeparc}. Analogous to how traditional software evolves to meet shifting requirements and enhance functionality, effective maintenance of DNNs is crucial for combating model drift and ensuring sustained accuracy and reliability as they are continuously exposed to new and changing data. 

Unlike traditional software systems characterized by their modular and discrete structures, DNNs are typically designed as cohesive, monolithic entities. The unique architecture introduces distinct and challenging maintenance tasks, including \textbf{model restructure}, \textbf{re-adaptation}, and \textbf{incremental development}. They are pivotal elements for the operational robustness and longevity of neural network-based systems in a rapidly advancing technological milieu~\cite{maintenance3}. By framing DNN maintenance as an extension of traditional software maintenance practices, we address a critical need within the software engineering community, ensuring that DNNs continue to function correctly, evolve smoothly, and remain aligned with evolving requirements.

Despite various proposals in the literature for maintaining DNN models like the DeepArc framework~\cite{ren2023deeparc}, these methods continue to face several limitations. One major challenge with traditional strategies, which depend largely on periodic model retraining, is their ineffectiveness in combating the unavoidable problem of model drift.
DeepArc addresses this problem by modularizing neural networks, organizing layers into modules based on their semantic similarities. This strategy streamlines maintenance tasks, facilitating the pruning and fine-tuning of specific modules or layers for more efficient updates.
However, this approach is constrained by its focus on layer-level manipulation. A more refined strategy that decomposes semantic components at a finer level, such as individual neurons, could enhance model restructuring and adaptation. Specifically, selectively pruning and tuning a small number of neurons might offer better effectiveness.
Furthermore, the incremental development of models~\cite{wang2024comprehensive} requires preserving the semantics and functions of previous versions while introducing new features. Given that semantics and functions in DNNs are primarily structured around neurons \cite{du2019techniques,xie2022npc,zhang2020interpreting,li2021understanding,shrikumar2017learning,li2024badedit,li2023fairer,runner}, the method of identifying and updating whole layers falls short of preserving the original semantics or functions during the incremental development process.

Acknowledging the existing limitations, our goal is to develop techniques that enhance DNN model maintenance. Drawing inspiration from the significant benefits of dynamic slicing in the maintenance and evolution of traditional software~\cite{beszedes2001dynamic,jiang2014accuracy}, we adopt similar slicing techniques for DNNs~\cite{Zhang2020Dynamic,Li2023Faire,xie2022npc,sun2022causality,liu2018fine}. Specifically, dynamic slicing excels in identifying and isolating essential code segments for implementing changes without causing unintended effects. By analogy, DNN slicing provides a systematic way to identify and separate semantic components at the neuron level, mirroring how code slicing pinpoints essential segments that influence a program’s behavior. Focusing interventions on the neurons most critical to the model’s functionality ensures targeted changes without introducing unintended side effects in non-critical parts.

\textbf{Our work.} We introduce the  \tool{} framework, building on the concept of semantic decomposition through DNN slicing, to innovate model maintenance practices. This framework features two main components: \textit{semantic slicing} and \textit{semantic-aware model maintenance}.  The former facilitates the precise identification of semantic elements within DNNs, such as critical neurons, at a granular level. Following this, the latter delivers tailored strategies for model maintenance tasks that demand an in-depth semantic comprehension, enabling effective and targeted strategies. 

The \textit{semantic slicing} module is designed to accurately identify critical neurons across the entire search space by employing a category-specific\footnote{To be precise, we identify critical neurons within the network model that have a significant impact on the output of each output neuron, where different categories refer to different output neurons. For example, in classification tasks, categories can be determined based on the labels in the dataset; in regression tasks, it can be considered a single-category task; in other types of tasks, different output neurons are treated as different categories.} strategy for optimal performance. Specifically, we compute \textit{neuron contribution scores} \cite{pei2017deepxplore,ish2019interpreting,xie2022npc,Li2023Faire,Molchanov2019Importance} for each neuron category individually, selecting a specific range of critical neurons for each category before combining them across the entire dataset.
Given the potential variability in the optimal range of critical neurons across different categories and layers—which may hinder effective combination—we refine this process to focus on neurons that encapsulate the layer's full semantic range. This is achieved by assessing the similarity between the characteristics of these chosen neurons and the entire layer. The subsequent merging strategy further enhances the utility and relevance of the identified critical neurons.

Mirroring and inspired by the traditional software engineering tasks, the \textit{semantic-aware model maintenance} module utilizes semantic slicing, and provides the following maintenance tasks for DNN models: 

\zsd{\ding{182} \textbf{Model Restructure}: In traditional software maintenance, code slicing can be used to identify redundant or non-essential code, streamlining the program and improving efficiency~\cite{10.1007/3-540-47764-0_3}. Similarly, in DNN model maintenance, model restructure focuses on retaining only the critical neurons necessary for maintaining predictive accuracy, while minimizing storage requirements. This step isolates non-essential components of the model, simplifying the overall structure by focusing on the critical parts.

\ding{183} \textbf{Model Re-adaptation}: Code slicing in software engineering helps identify and modify the core parts of a program when fixing bugs~\cite{10.1145/3579640, 10.1109/ASE56229.2023.00073,10.1145/3639478.3643117}. In DNNs, model re-adaptation focuses on retraining critical neurons to address prediction errors, reducing the need to adjust many parameters. This mirrors how targeted changes in code slicing resolve issues without overhauling the entire program.

\ding{184} \textbf{Model Incremental Development}: In traditional software maintenance, slicing can be used to store the historical semantic information of code, thereby assisting software evolution~\cite{10.1145/2393596.2393646, 8115722}. Similarly, model incremental development stabilizes critical neuron parameters to prevent catastrophic forgetting when integrating new knowledge. This ensures that core logic remains intact while new functionality is added seamlessly, akin to how code slicing maintains program stability during incremental updates.}

We conduct extensive experiments to evaluate the performance of \tool{}. For model restructure, \tool{} enhances the efficiency of existing model compression techniques by an average of 13.46\%, while also improving compression rate. For model re-adaptation, \tool{} only trains an average of 61.49\% of the parameters, enhancing both accuracy and efficiency. For incremental model development, \tool{} performs well, boosting average accuracy by approximately 40\% compared to direct retraining.

\textbf{Contributions.} The contributions of this paper is summarized as follows: \ding{182} We propose \tool{}, including \textit{semantic slicing} for finer-grained neuron-level decomposition, and \textit{semantic-aware model maintenance}, offering novel and effective strategies for semantic-aware DNN model maintenance tasks. \ding{183} We detail a sophisticated \textit{semantic slicing} technique that employs a category-specific strategy to accurately identify and combine critical neurons across the dataset, refining the process of neuron selection based on semantic range to optimize model performance. \ding{184} We tackle the challenging maintenance tasks via the \textit{semantic-aware model maintenance} module, which significantly advances the tasks of \textit{model restructure}, \textit{re-adaptation}, and \textit{incremental development}.  \ding{185} We conduct extensive experiments on three model maintenance tasks, showcasing the superior performance of \tool{}.

%% file: Chapters/background.tex
\section{Background and Related Work}
\subsection{Deep Neural Networks}
A DNN typically consists of multiple layers of neurons \cite{tishby2015deep}, which can be formally defined as follows.

\begin{definition}
A Deep Neural Network (DNN) $f$ consists of $L$ multiple layers $\left\langle l_0, l_1,\ldots,l_{L-1}\right\rangle$, where $l_0$ is the input layer, $l_{L-1}$ is the output layer, and $l_1,\ldots,l_{L-2}$ are hidden layers. The inputs of each layer are the outputs of the previous layer. 
\end{definition}

In this work, we mainly focus on the classifier $f:\mathcal{X} \rightarrow \mathcal{Y} $, where $\mathcal{X}$ is a set of inputs and $\mathcal{Y}$ is a set of classes. Given an input $x\in \mathcal{X}$, 
we use $f^l(x)$ to represent the internal features extracted by the layer $l$ and $f_n(x)$ to represent the internal features extracted by the neuron $n$ (\ie, the output values of neurons at $l$). $\hat{y}$ represents the output of DNN (\ie, $f(x)$) and $y$ is the label of the given input $x$.

There usually exists redundancy in DNNs which refers to components within the network that do not significantly contribute to its performance \cite{cheng2015exploration,xie2022npc}. 
This redundancy can manifest in various forms, such as redundant neurons and redundant layers.
Nevertheless, determining which components are redundant and can be removed without impacting performance requires sophisticated analysis and careful validation.

\subsection{DNN Model Maintenance}
\label{sec:aimm}
In the realm of software engineering, maintaining and evolving systems has long been recognized as a critical challenge, and DNNs are no exception. As these models increasingly integrate into complex, safety-critical, and rapidly changing environments, practitioners must apply well-established maintenance principles—such as iterative improvement, modular refactoring, and continuous quality assurance—to keep DNNs robust, adaptable, and high-performing. This concept, akin to the established practices of traditional software maintenance, addresses the evolving needs within the software engineering domain for more efficient and adaptable DNN systems.

DeepArc~\cite{ren2023deeparc} has introduced a new neural network architecture strategy to cut model maintenance costs. This approach identifies semantically similar layers as redundant, allowing for model restructuring by pruning these layers. Only non-redundant layers are fine-tuned on the target data, based on calculated semantic similarities of layer representations:
Given a dataset $D$, the matrix $M^{l_i}$ is denoted as the similarity matrix for the $i$-th network layer $l_i$, where 
\begin{align}
\label{eq:matrix}
M^{l_i}[a,b]= f^{l_i}(x_a) \cdot f^{l_i}(x_b),
\end{align}
with $x_a$ and $x_b$ representing any two inputs from the dataset $D$. 
Intuitively, $M^{l_i}$ indicates the embedding landscape of the dataset on the $i$-th layer and the shape of the matrix is solely dependent on the size of the dataset. 
Similar embedding landscapes (\ie, similar matrices) in two layers indicate they extract comparable semantic information from inputs.
With the matrix $M^{l}$ for each layer $l$, DeepArc then follows the centered kernel alignment \cite{kornblith2019similarity} (CKA) metric to calculate the similarity between $M^{l_i}$ and $M^{l_j}$ to represent the semantic similarity between layer $f^{l_i}$ and $f^{l_j}$.

\begin{align}
   \text{sim}_{\text{sem}}(f^{l_i},f^{l_j}) =  \text{CKA}(M^{l_i},M^{ l_{j}}) &= \frac{M^{l_i} \cdot M^{l_{j}}} {||M^{l_i}|| \cdot || M^{ l_{j}}||}. \label{eq:sim}
\end{align}

A higher $\text{sim}_{\text{sem}}$ indicates greater semantic similarity between consecutive layers, indicating redundancy. Removing these redundant layers aids in model restructuring, while focusing retraining efforts on only the essential, non-redundant layers facilitates model re-adaptation. This strategy, however, only targets redundant layers without considering redundant neurons, causing limitations to be detailed in Section~\ref{sec:m1}.

\subsection{DNNs Slicing}
\label{sec:slicing}
DNNs, essentially programs made up of artificial neurons, have seen recent innovations in ``slicing'' techniques. \zsd{These techniques, although termed as ``slicing,'' essentially represent advanced methods for \textbf{neuron selection}. The key idea behind DNN slicing is to evaluate and identify the most influential neurons in a network for various tasks~\cite{xie2022npc,Zhang2020Dynamic}. In this context, DNN slicing is closely related to \emph{neuron contribution metrics}, which quantify the impact of each individual neuron on the overall performance of the model, particularly on a dataset $D$. The essence of DNN slicing research is not only to partition the network but to systematically select neurons that significantly contribute to the network’s functionality, based on their contribution scores. A \emph{neuron selection strategy} is then applied to identify and extract these critical neurons, thereby enabling focused modifications, optimizations, or analysis of the network's performance.}

Specifically, the process could be formulated as follows:
we calculate $\mathbf{C}_n^{l,D}$ for each neuron $n$ in layer $l$ based on a dataset $D$. The \emph{neuron contribution} of the layer $l$ composed of neuron set $\mathbf{N}^l$ could be represented as $\mathbf{C}^{l,D} = \{\mathbf{C}_{n_0}^{l,D}, \mathbf{C}_{n_1}^{l,D}, ..., \mathbf{C}_{n_{|\mathbf{N}^l|-1}}^{l,D}\}$. 
We further determine the \emph{selection} of critical neurons according to the calculated neuron contribution and denote the selected critical neurons as $\mathbf{\Omega}^{l,D}_k$:
\begin{equation}
\label{eq:omega}
    \mathbf{\Omega}_k^{l,D} = \text{top}\-k(\mathbf{N}^{l}, \mathbf{C}^{l,D})
\end{equation}
where $\text{top}\-k (\cdot)$ represents the top $k$ maximum instances of the input set $\mathbf{N}^l$ based on certain metrics. 
We can further depict the critical neurons of the model $f$ as $\mathbf{\Omega}_k^{D} = \{\mathbf{\Omega}_k^{l_0,D}, \mathbf{\Omega}_k^{l_1,D}, ..., \mathbf{\Omega}_k^{l_{L-1},D}\}$. Note the proportion of selected critical neurons is directly controlled by hyperparameter $k$.

The predominant approaches of \textit{neuron contribution metric} design for neuron $n$ are as follows:
\ding{182} \textbf{Avg} (Average of Neuron Activation \cite{pei2017deepxplore}). We compute the mean activation value for each neuron across the dataset, defined as: $avg_{n}(D)=\frac{\sum_{x \in D} f_{n}(x)}{|D|}.$ 
\ding{183} \textbf{Var} (Variance of Neuron Activation \cite{ish2019interpreting}). We compute the activation value variance for each neuron using the formula: $var_{n}(D)=\frac{\sum_{x \in D}(f_{n}(x)- \overline{ f_{n}(x)})^2}{|D|}.$
\ding{184} \textbf{Ens} (Ensemble of Neuron Activation). For each neuron, we compute its $avg_n(D)$ and $var_n(D)$ on dataset $D$, normalize both values to [0,1], and then combine them into a single ensemble-based contribution score.
\ding{185} \textbf{DeepLIFT} (Interpretability-based Score \cite{xie2022npc,Li2023Faire}). Given its interpretability, our analysis predominantly utilizes DeepLIFT~\cite{shrikumar2017learning} for computations.
Specifically, DeepLIFT assigns contribution scores to each neuron based on the \emph{gradient} during the backward propagation process.
\ding{186} \textbf{Taylor} (Pruning-oriented Score~\cite{Molchanov2019Importance}).
 Taylor utilizes \emph{model weights} and \emph{gradient information} to assess the contribution of neurons.

Leveraging the current neuron contribution metrics, there are intuitively two neuron selection strategies. The \emph{global neuron selection method} calculates neuron contribution scores across the entire dataset $D$, followed by a selection of a specific range of neurons. In contrast, the \emph{category-wise neuron selection approach} computes scores for each category of data ($D_{c_0}$, $D_{c_1}$, $\ldots$) separately, identifies a range of critical neurons for each category (\ie, identifying $\mathbf{\Omega}_k^{l,D_{c_0}}, \mathbf{\Omega}_k^{l,D_{c_1}}, \ldots$), and ultimately merges these category-wise critical neurons for the entire dataset as a whole ($\mathbf{\Omega}_k^{l,D} $).

%% file: Chapters/motivation.tex
\begin{figure*}
    \centering
    \includegraphics[width=0.95\textwidth]{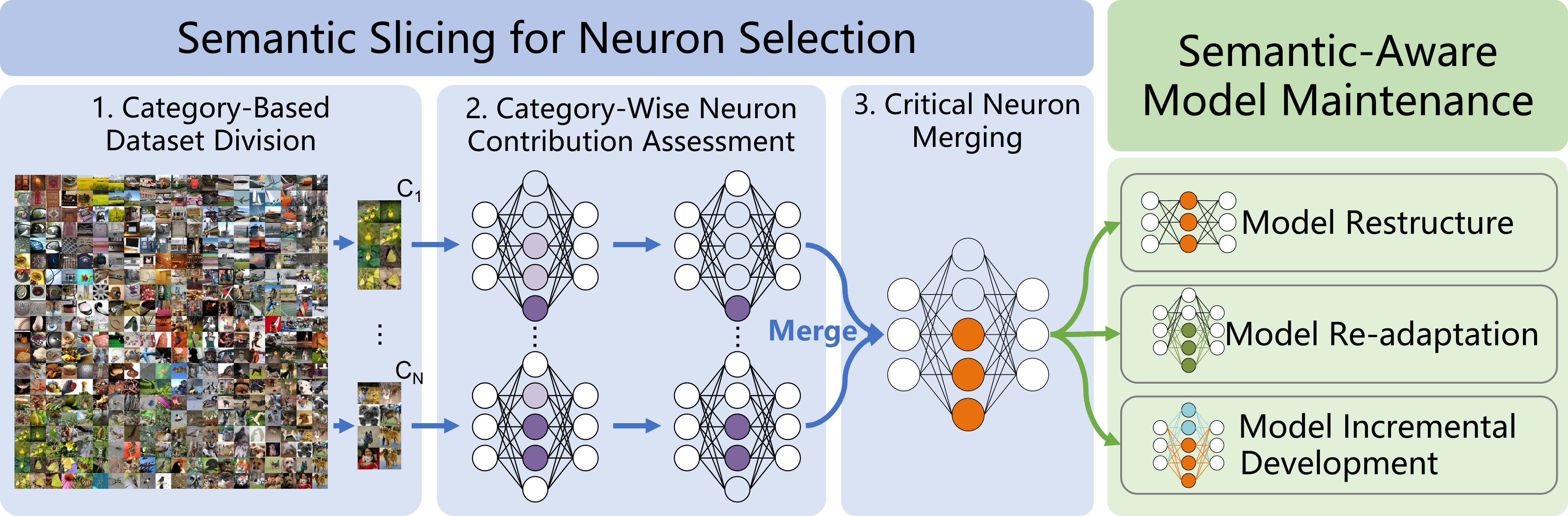}
    \caption{Overview of \tool{} Framework. In Semantic Slicing, Neuron Contributions Are Highlighted in \textcolor{darkpurple}{$purple$}, With Darker Shades Signifying Greater Contributions. Critical Neurons Are Indicated in \textcolor{orange}{$orange$}. For Model Restructure, We Preserve Only These Critical Neurons. During Model Re-Adaptation, We Train Solely the Critical Neurons, Marked in \textcolor{darkgreen}{$green$}. For Model Incremental Development, Only the Non-Critical Neurons Are Trained, Shown in \textcolor{lightblue}{$blue$}.}
    \label{fig:Method_Overview}
    \vspace{-10pt}
\end{figure*}

\subsection{Motivation}
\label{sec:m1}
\zsd{Maintaining DNN models presents significant challenges, especially as both models and datasets continue to scale in size and complexity. Current state-of-the-art techniques~\cite{ren2023deeparc} exhibit several limitations that impede effective model maintenance:

\ding{182} \textbf{Insufficient Granularity.} Existing methods predominantly operate at the layer level, which restricts their ability to perform precise model restructuring and adaptation. This coarse granularity overlooks critical details at the neuron level, where fine-grained adjustments to individual neurons—especially those supporting specific functionalities—can significantly enhance model performance and efficiency.

\ding{183} \textbf{Ineffectiveness for Relatively Small Models.} The term ``small models'' in the context of this paper often refers to models that are relatively small compared to the size of their training datasets, rather than being absolutely small. DeepArc~\cite{ren2023deeparc} has demonstrated that layer-level modularization is most effective when the model's scale significantly exceeds that of the dataset. For instance, DeepArc's experiments utilized large models such as ResNet-110, Wide-ResNet-38, and VGGNet-19 trained on relatively small datasets like MNIST, FMNIST, and CIFAR-10. Our own experiments (RQ1) further confirm that when the disparity between model size and dataset size is reduced, the benefits of layer-level modularization diminish or disappear entirely. This indicates that layer-level maintenance techniques are less effective for models that are not excessively large relative to their training data. Consequently, there is a need for more refined, neuron-level approaches to ensure efficient and scalable model maintenance across varying model and dataset scales.

\ding{184} \textbf{Lack of Incremental Model Development.} Incremental model development is a crucial aspect of model maintenance that remains inadequately addressed by current techniques. Existing approaches struggle to identify and preserve key components at the neuron level~\cite{du2019techniques}, which are essential for maintaining the model's functionality. This limitation poses significant challenges in developing new features or functions while ensuring that existing capabilities are retained, especially when relying solely on layer-level adjustments. A neuron-level maintenance strategy would enable more precise and incremental updates, facilitating the continuous evolution of DNN models without compromising their existing performance.

Addressing these limitations requires a shift towards more granular maintenance methodologies. By adopting neuron-level strategies, it becomes possible to achieve finer control over model restructuring and adaptation, enhance the effectiveness of maintenance for models of varying sizes relative to their datasets, and support incremental development processes. This approach promises to provide a more flexible and robust framework for DNN model maintenance, ultimately leading to more efficient and scalable deployment of deep learning models in diverse applications within the field of software engineering.
}

\subsection{\tool{} Overview}
To address the issues, we propose \tool{} which focuses on neuron-level DNN semantic decomposition and involves slicing DNNs to identify and isolate critical neuron components, facilitating a variety of model maintenance tasks. 
Figure \ref{fig:Method_Overview} shows a detailed workflow of our \tool{} framework, which mainly consists of two modules: \textit{Semantic Slicing} and \textit{Semantic-aware Model Maintenance}.

\textit{Semantic slicing.} This module initially identifies and extracts crucial neurons across the entire model. These neurons are representative of the semantics inherent in each layer of the model. 
By focusing on a more detailed level, it overcomes the initial two limitations, enabling more precise restructuring and adaptation. Additionally, it improves the ability to identify redundant neurons in the smaller layers of models.

\textit{Semantic-aware model maintenance.} By utilizing the aggregated critical neurons in the model, this approach supports various applications, offering more robust solutions for long-term DNN model maintenance. Beyond restructuring and re-adapting the model, this method also addresses incremental model development. It specifically counteracts catastrophic forgetting, which occurs when an AI model learns new information, by stabilizing the critical neurons' parameters. This enables the model to incrementally develop functions with limited access to the original training data.

In Section \ref{sec:sl}, we will provide a detailed introduction to the design methodology behind \textit{Semantic Slicing}. Following that, in Section \ref{sec:mmt}, we will discuss \textit{Semantic-aware Model Maintenance}: the application of \textit{Semantic Slicing} in the context of model maintenance tasks.

\section{Semantic Slicing}
\label{sec:sl}
As discussed in Section~\ref{sec:slicing}, various neuron contribution metrics and two possible neuron selection strategies can be adopted for neuron-level DNNs slicing. Yet, how well these methods perform in model maintenance tasks has not been fully examined. To address this, Section \ref{sec:existing_setting} details a pilot study designed to evaluate their effectiveness and efficiency in maintaining models. Following these experiments, we present our approach to semantic slicing in Section \ref{sec:semslicing}, guided by the insights gained.

\subsection{A Pilot Study on Existing DNNs Slicing Strategies}
\label{sec:existing_setting}

\noindent\textbf{Global Slicing.}
One possible strategy is the global approach, which calculates contribution metrics across the entire dataset and ranks all neurons accordingly. This process, outlined in Section \ref{sec:slicing}, results in five distinct sets of critical neurons identified using different metrics. We then group neurons into different percentile ranges (\eg, top 0–10\%, 10–20\%, etc.) and devise a strategy to evaluate the importance of each set to determine the most effective contribution metric.
Specifically, to assess and compare the significance of the critical neurons identified by various metrics, we adopt the method from Xie et al.~\cite{xie2022npc} by zeroing the outputs of these neurons and observing the impact on model predictions. Greater changes in predictions suggest higher neuron significance.

The study is conducted on CIFAR-10 \cite{krizhevsky2009learning}, using VGG-16 \cite{simonyan2014very} for classifying. Table \ref{tab:Global_Settings} shows that in a global setting, all neuron contribution metrics have limitations. Surprisingly, masking neurons deemed less critical often leads to more significant accuracy losses, suggesting a misalignment between globally calculated contributions and the neurons' true importance. For example, using the Avg metric, masking 20\%$\rightarrow$30\% of neurons has a greater impact than masking the top 10\%$\rightarrow$20\%, as highlighted by the green arrows in the table. The results suggest that global slicing is not effective for model maintenance tasks.

\begin{table}[]
\centering
\caption{Model Accuracy After Masking Selected Critical Neurons Globally Calculated by Different Contribution Metrics. The Anticipated Trend in Model Accuracy Is an Increase (Signified by \textcolor{red}{$\uparrow$}), Correlating With the Diminishing Contribution of Masked Neurons. Nonetheless, Within the Context of Global Slicing, Numerous Anomalous Instances of Decreased Model Accuracy Are Observed, as Indicated by \textcolor{green}{$\downarrow$}.}
\vspace{-0.2cm}
\label{tab:Global_Settings}
\resizebox{0.8\columnwidth}{!}{
\begin{tabular}{cccccc}
\hline
\multicolumn{6}{c}{VGG16-CIFAR10} \\ \hline

\begin{tabular}[c]{@{}c@{}}performance\\ after masking \end{tabular} & Avg $\textcolor{red}{\uparrow}$ & Var $\textcolor{red}{\uparrow}$ & Ens $\textcolor{red}{\uparrow}$ & DeepLIFT $\textcolor{red}{\uparrow}$ & Taylor $\textcolor{red}{\uparrow}$ \\ \hline
0\%-\textgreater{}10\% & 42.85\% & 46.78\% & 50.64\% & 25.78\% & 41.29\% \\
10\%-\textgreater{}20\% & 65.45\% & 76.91\% & 62.27\% & 36.83\% & 47.94\% \\
20\%-\textgreater{}30\% & {60.32\%$\textcolor{green}{\downarrow}$} & {71.28\%$\textcolor{green}{\downarrow}$} & {57.82\%$\textcolor{green}{\downarrow}$} & {35.49\%$\textcolor{green}{\downarrow}$} & 60.13\% \\
30\%-\textgreater{}40\% & 69.15\% & {69.53\%$\textcolor{green}{\downarrow}$} & 70.53\% & 65.94\% & 64.67\% \\
40\%-\textgreater{}50\% & {65.34\%$\textcolor{green}{\downarrow}$} & {49.47\%$\textcolor{green}{\downarrow}$} & { 62.55\%$\textcolor{green}{\downarrow}$} & 63.84\%$\textcolor{green}{\downarrow}$ & 72.05\% \\
\hline
\end{tabular}
}\vspace{-0.4cm}
\end{table}

\begin{table}[]
\centering
\caption{The Model Accuracy Averaged by Category After Masking Critical Neurons Derived by Various Metrics in Categorical Settings. The Anticipated Trend in Model Accuracy Is an Increase (Signified by \textcolor{red}{$\uparrow$}), Correlating With the Diminishing Contribution of Masked Neurons.}
\vspace{-0.2cm}
\label{tab:Categorical_Settings}
\resizebox{0.8\columnwidth}{!}{
\begin{tabular}{cccccc}
\hline
\multicolumn{6}{c}{VGG16-CIFAR10} \\ \hline
\begin{tabular}[c]{@{}c@{}}performance\\ after masking\end{tabular} & \multicolumn{1}{c}{Avg $\textcolor{red}{\uparrow}$} & \multicolumn{1}{c}{Var $\textcolor{red}{\uparrow}$} & Ens $\textcolor{red}{\uparrow}$ & DeepLIFT $\textcolor{red}{\uparrow}$ & \multicolumn{1}{c}{Taylor $\textcolor{red}{\uparrow}$} \\ \hline
0\%-\textgreater{}10\% & 0.00\% & 0.09\% & 0.00\% & 0.30\% & 11.80\% \\
10\%-\textgreater{}20\% & 0.07\% & 5.16\% & 0.12\% & 8.48\% & 27.94\% \\
20\%-\textgreater{}30\% & 1.25\% & 11.02\% & 2.49\% & 21.49\% & 48.24\% \\
30\%-\textgreater{}40\% & 31.39\% & 37.81\% & 19.58\% & 59.12\% & 53.90\% \\
40\%-\textgreater{}50\% & 55.45\% & 58.63\% & 57.61\% & 80.79\% & 62.64\% \\
\hline
\end{tabular}
}\vspace{-13px}
\end{table}

\noindent\textbf{Category-wise Slicing.}
Another possible strategy is the category-wise approach. Acknowledging that neuron contributions can vary widely between categories, which may limit the effectiveness of global DNN slicing, we conduct a category-specific analysis. We calculate the five metrics for each category and then conduct masking experiments on data from each category (\ie, zeroing the outputs of neurons deemed critical for that category) to assess accuracy on a per-category basis. The averaged and individual category results are shown in Tables \ref{tab:Categorical_Settings} and \ref{tab:class_results}, respectively.
Masking the top 10\% of neurons shows that certain metrics~(\eg, Avg and Ens) drastically lowered classification performance, nearly to zero. This demonstrates that a category-specific approach is more effective in identifying critical neurons than that of the global-based approach.

\subsection{Identified Challenges for Semantic Slicing}
Identifying critical neurons for model maintenance involves two main challenges: the optimal proportion of critical neurons varies by category and by layer. \ding{182} \textbf{Different category-wise distribution.} Integrating category-specific critical neurons into a set for the entire dataset is complex as the ideal proportion of these neurons differs across categories. For example, masking 10\% to 20\% of critical neurons in Category 4 hardly affects performance, suggesting redundancy. In comparison, masking 30\% to 40\% in Category 1 significantly impacts performance, indicating their importance for that category, as shown in Table~\ref{tab:class_results}. \ding{183} \textbf{Different layer-wise distribution.} The distribution of critical neurons changes across layers, often becoming sparser in higher layers. Current methods propose a uniform selection range for all categories and layers, which can either overlook crucial neurons or include too many non-critical ones, affecting accuracy and efficiency in model maintenance. Therefore, determining the optimal range of critical neurons for each category and layer is essential for effective DNN slicing.

However, achieving the goal is a challenging endeavor. The challenge lies in the impracticality of conducting an exhaustive search across all possible combinations of critical neuron proportions for each layer and category to assess performance after merging. Both the exhaustive search and the post-merging evaluation are inefficient in their current forms.
To address this, we propose \textit{Semantic Slicing}, to be detailed next in Section~\ref{sec:semslicing}.

\begin{table}[]
\centering
\caption{Model Performance on Selected Categories After Masking Selected Critical Neurons Calculated by Different Contribution Metrics. The Anticipated Trend in Model Accuracy Is an Increase (Signified by \textcolor{red}{$\uparrow$}), Correlating With the Diminishing Contribution of Masked Neurons.}
\vspace{-0.2cm}
\label{tab:class_results}
\resizebox{0.8\columnwidth}{!}{
\begin{tabular}{ccccc}
\hline
\multicolumn{5}{c}{VGG16-CIFAR10 (DeepLIFT)} \\ \hline
\multicolumn{1}{c}{} & \multicolumn{1}{c}{Category 1 $\textcolor{red}{\uparrow}$} & \multicolumn{1}{c}{Category 4$ \textcolor{red}{\uparrow}$} & \multicolumn{1}{c}{Category 7 $\textcolor{red}{\uparrow}$} & \multicolumn{1}{c}{Category 10 $\textcolor{red}{\uparrow}$} \\ \hline
0\%-\textgreater{}10\% & 0.00\% & 3.00\% & 0.00\% & 0.00\% \\
10\%-\textgreater{}20\% & 0.00\% & 81.80\% & 0.00\% & 0.00\% \\
20\%-\textgreater{}30\% & 13.30\% & 80.40\% & 31.70\% & 5.90\% \\
30\%-\textgreater{}40\% & 33.20\% & 90.20\% & 39.20\% & 86.70\% \\
40\%-\textgreater{}50\% & 78.50\% & 95.80\% & 90.80\% & 96.30\% \\
\hline
\end{tabular}
}\vspace{-0.5cm}
\end{table}

\subsection{The Design of Semantic Slicing}
\label{sec:semslicing}

The essence of existing DNNs slicing methods revolves around employing a contribution metric and then taking a neuron selection strategy to determine the range of critical neurons. Compared with these approaches, the key novelty of semantic slicing lies in the semantic-based neuron selection strategy. 
The aim of semantic slicing is to address the issue of determining the appropriate $k$ value in Equation \ref{eq:omega} for various layers and categories.
To address this issue, we here recognize that the crux of choosing critical neurons lies in \textit{selecting those neurons whose semantics are adequately representative of the entire layer}.

\noindent\textbf{Problem Definition.}
Inspired by prior layer similarity evaluation works~\cite{nguyen2020wide,ren2023deeparc}, 
we formally reformulate the task of selecting top $k$ neurons for each layer and category as follows. For each layer $l$ and category $c$, identify $k$ neurons such that they fulfill the condition:
$\text{sim}_{\text{sem}}( f^{\mathbf{\Omega}_k^{l,c}}, f^{l,c}) > \Theta$, where $\Theta$ represents a unified threshold applicable across various categories and layers, and $\Omega_k^{l,c}$ is the selected neuron set.
Employing the CKA metric as in Equation \ref{eq:sim}, we calculate a similarity metric in the range of [0, 1] as follows:
\vspace{-0.2cm}
\begin{align}
 min |\mathbf{\Omega}^{l,c}| \quad \text{subject to} \quad \frac{M^{\mathbf{\Omega}^{l,c}} \cdot M^{l,c}} {||M^{\mathbf{\Omega}^{l,c}}|| \cdot || M^{l,c}||} > \Theta, \label{eq:sim2}\vspace{-0.4cm}
\end{align}
where $\mathbf{\Omega}^{l,c}$ is a neurons subset of layer $l$ selected on category $c$, and $M^{l,c}$ is the embedding landscape calculated on category $c$ according to Equation \ref{eq:matrix}. Given the neuron contribution set $\mathbf{C}^{l,c} = \{\mathbf{C}_{n_0}^{l,c}, \mathbf{C}_{n_1}^{l,c}, ..., \mathbf{C}_{n_{|\mathbf{N}^l|-1}}^{l,c}\}$, we can further reformulate the given problem into a variant of the \textit{Prefix Sum Problem} \footnote{
The Prefix Sum problem, in its simplest form, involves computing an array of cumulative sums (prefix sums) of a given sequence of numbers.}. Specifically, the objective is to identify a particular value of $k$ such that $\text{sim}_{\text{sem}}( f^{\mathbf{\Omega}_k^{l,c}}, f^{l,c}) > \Theta$ given the layer-wise contribution value set $\mathbf{C}^{{l,c}}$.

\zsd{
\noindent\textbf{Algorithm Implementation.} To address the aforementioned issue, we adopt a \textit{Linear Scan} strategy as described by Algorithm~\ref{alg}. We break down the implementation into three steps:

\begin{enumerate}
    \item \textbf{Initialization:}
    We begin by selecting the top $i$ neurons with the highest contribution scores from $\mathbf{C}^{l,c}$ to form the initial set $\mathbf{\Omega}^{l,c}$. At this stage, $\mathbf{\Omega}^{l,c}$ contains a small but promising subset of neurons for category $c$ at layer $l$. Using this initial set, we compute $\mathrm{sim}_{\mathrm{sem}}\bigl(f^{\mathbf{\Omega}^{l,c}}, f^{l,c}\bigr)$ to assess how well it captures the semantic characteristics of the entire neuron population $\mathbf{N}^l$.

    \item \textbf{Iteration Until Threshold:}
    If the current similarity does not surpass the predefined threshold $\Theta$, we continue the selection process. This involves identifying the next top $i$ neurons from the remaining pool $\mathbf{N}^l \setminus \mathbf{\Omega}^{l,c}$ according to their contribution scores. By iteratively adding these high-contribution neurons, we gradually refine the subset $\mathbf{\Omega}^{l,c}$, ensuring that it becomes more semantically aligned with $f^{l,c}$.

    \item \textbf{Stopping Condition:}
    The iteration proceeds until we achieve $\mathrm{sim}_{\mathrm{sem}}\bigl(f^{\mathbf{\Omega}^{l,c}}, f^{l,c}\bigr) > \Theta$. Once the threshold is met, the algorithm terminates, and the resulting set $\mathbf{\Omega}^{l,c}$ is deemed critical for capturing the semantic features of category $c$ at layer $l$. To further reduce the computational load, practitioners may choose a larger increment $i$, thereby requiring fewer iterations and faster convergence.
\end{enumerate}

\begin{algorithm}[t]
\small
\caption{Linear Scan Algorithm.}
\label{alg}
\KwData{All neurons $\mathbf{N}^l$ of layer $l$, the neuron contribution set $\mathbf{C}^{l,c}$, the semantic similarity threshold $\Theta$, the selection interval $i$.}
\KwResult{The critical neuron set for category $c$ at layer $l$: $\mathbf{\Omega}^{l,c}$.}

$\Omega^{l,c} \leftarrow \text{top } i (\mathbf{N}^l,\mathbf{C}^{l,c})$

\While{$ \mathrm{sim}_{\mathrm{sem}}\bigl(f^{\Omega^{l,c}},f^{l,c}\bigr) < \Theta$}{
    $\Omega^{l,c} \leftarrow \Omega^{l,c} \cup \text{top } i (\mathbf{N}^l \setminus \Omega^{l,c}, \mathbf{C}^{l,c})$
}
\end{algorithm}}

\noindent\textbf{Categorical Critical Neurons Aggregation.}
After determining the critical neurons for each category, they can be combined on a category basis to establish the critical neuron set for the entire model, as shown in Figure \ref{fig:Method_Overview}. For an entire dataset $D$ containing $m$ categories, this can be expressed as: $\mathbf{\Omega}^{l,D} = \mathbf{\Omega}^{l,D_{c_0}} \cup \mathbf{\Omega}^{l,D_{c_1}} \cup \ldots \cup \mathbf{\Omega}^{l,D_{c_{m-1}}} $. We can further represent the critical neuron set of the model $f$ as $\mathbf{\Omega} = \{\mathbf{\Omega}^{l_{0,D}}, \mathbf{\Omega}^{l_{1,D}}, ..., \mathbf{\Omega}^{l_{L-1,D}}\}$. 
Given the neuron contribution metrics \(\mathbf{C}\), the selection threshold \(\Theta\) directly determines \(\mathbf{\Omega}\), simplifying the process by eliminating the need to consider varying optimal neuron proportions across categories and layers.

%% file: Chapters/methodology.tex
\setlength{\parindent}{2em}
\section{Semantic-Aware Model Maintenance}
\label{sec:mmt}
\subsection{Model Restructure}
\begin{figure*}
    \centering
    \includegraphics[width=0.8\textwidth]{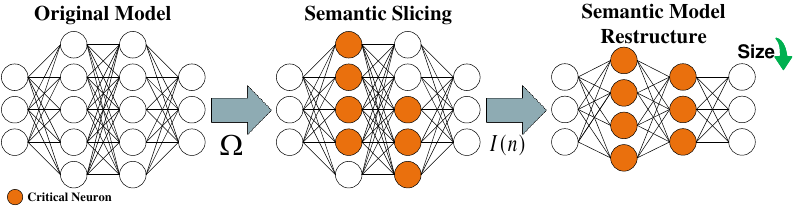}
    \caption{Model Restructure Based on Semantic Slicing. We Only Retain the Critical Neurons for Model Restructure.}
    \label{fig:Restructure}
    \vspace{-10pt}
\end{figure*}

Identifying critical neurons enables the removal of non-critical ones, keeping only essential neurons in each layer. This reduces the model's size by eliminating `redundant' elements while preserving prediction accuracy, as shown in Figure~\ref{fig:Restructure}. This process uses an indicator function $I(n)$ for each neuron $n$ to decide whether to retain it based on its contribution. 
The model restructure strategy can be expressed as:\vspace{-0.1cm}
\begin{align}
\label{eq:Restructure}
I(n) = \left\{ \begin{array}{ll}
1 & \text{if } n \in \mathbf{\Omega}, \\
0 & \text{otherwise}.
\end{array} \right.
\end{align}
Then each neuron $n'$ in the restructured model $f'$ could be expressed as $n'=n \cdot I(n)$. Note that the threshold value \(\Theta\) can be dynamically adjusted to balance model performance and size during restructuring. By adjusting \(\Theta\), this method allows for a restructured model that varies in accuracy and size to suit diverse needs. Such restructuring will lead to an effective decrease in the storage space. 

\zsd{Moreover, some existing pruning techniques (e.g.,~\cite{Lee_Park_Mo_Ahn_Shin_2020, Morcos_Yu_Paganini_Tian_2019, Gale_Elsen_Hooker_2019, Evci_Gale_Menick_Castro_Elsen_2019}) primarily operate at the weight level using unstructured approaches. These methods rely on approximate calculations to reduce weights while maintaining approximate model performance. In contrast, our proposed model restructure method identifies semantic slices (i.e., critical neurons) at the neuron level and removes non-critical neurons to compress the model. By leveraging model restructure, our method effectively complements existing pruning techniques, enabling these weight-based pruning methods to focus on redundant weights within critical neurons, thereby accelerating the pruning process and achieving better model compression.}

\subsection{Model Re-adaptation}
Model re-adaptation improves adaptability and robustness by focusing on training parameters related to critical neurons. 
Previous research~\cite{xie2022npc,zhang2020interpreting,li2021understanding} has shown that in cases of poor performance, such as adversarial examples, adversarial noise is amplified and propagated through critical neurons. Therefore, the fundamental cause is the abnormal behavior of critical neurons leading to errors, while other neurons are less affected.
Retraining these critical neurons' parameters can thus more efficiently repair and enhance the model in terms of cases with poor performance.
The complete model re-adaptation workflow is shown in Figure \ref{fig:Re-adaption}.

\zsd{
Using adversarial samples as an example, the retraining strategy explicitly incorporates both clean and adversarial inputs into the training objective. By including adversarial examples $(adv_x, y)$ alongside their benign counterparts $(x, y)$, the retraining process directly addresses the model's susceptibility to adversarial perturbations. Concretely, the loss function is defined as:
\begin{align}
\label{eq:ReadaptionLoss}
\mathcal{L} = \mathcal{L}_{c}(x, y; \theta) + \mathcal{L}_{c}(adv_x, y; \theta),
\end{align}
where $L_{c}$ denotes the cross entropy, $adv_x$ is the adversarial counterpart of the sample $x$, and the notation $\theta$ represents the parameters in model $f$. By summing the losses from both clean and adversarial samples, this objective ensures that the critical neurons learn to correctly classify benign inputs while simultaneously increasing their resistance to adversarial noise. The inclusion of adversarial data in the objective effectively provides stronger gradient signals focused on those regions of the parameter space associated with vulnerability, guiding critical neurons to adjust their representations.

To implement this targeted retraining, the parameter update strategy is carefully restricted. Instead of applying gradient-based updates to all parameters in the model, the procedure focuses exclusively on the parameters associated with the critical neuron set $\mathbf{\Omega}$. Formally, given a learning rate $\eta$, the parameter update rule is:
\begin{align}
\label{eq:ReadaptionParameters}
\theta'_{\mathbf{\Omega}} = \theta_{\mathbf{\Omega}} - \eta \cdot \nabla_{\theta_{\mathbf{\Omega}}} \mathcal{L},
\end{align} 
This selective update confines optimization to parameters identified as critical, effectively targeting the root causes of performance issues.

\begin{figure*}
    \centering
    \includegraphics[width=0.62\textwidth]{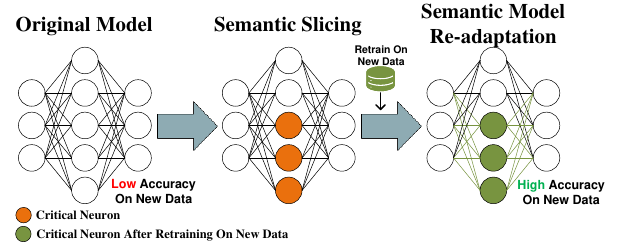}
    \caption{Model Re-Adaptation Based on Semantic Slicing. We Train Only the Critical Neurons for Model Re-Adaptation.}
    \label{fig:Re-adaption}
    \vspace{-15pt}
\end{figure*}

}
\subsection{Model Incremental Development}
Model incremental development refers to the ability of a DNN model to learn from new data over time without forgetting previously learned information. This is crucial for creating DNN models that can adapt and evolve in dynamic environments.
Taking an image classification designed to recognize different types of animals as an example, we initially only have data to train a DNN to classify images into three categories: Dogs, Cats, and Birds (old task).  After the initial training, we collect more data and want the DNN to learn to recognize two new categories Fish and Horses (new task).
This poses a significant challenge for DNN models: \ding{182} In practical applications, the training data for the old task might be discarded to save storage space and we only have access to the data of the new task. When training only with the data of the new task, the incremental development of DNNs is prone to ``catastrophic forgetting'', where learning new tasks can lead to the loss of old tasks.  \ding{183} Even with the training data of the old task, the training process is less resource-efficient because the DNN models have to be retrained with the entire dataset (new and old data).\looseness=-1

This issue is particularly prominent in real-world scenarios, such as in Internet of Things (IoT) devices~\cite{9425491}, where incremental development is essential. Due to space constraints, these devices might initially be trained on a limited set of data categories. However, as new data becomes available, it is often impractical to retrain the entire model from scratch or to delete previously learned categories.

\begin{figure*}
    \centering
    \includegraphics[width=0.58\textwidth]{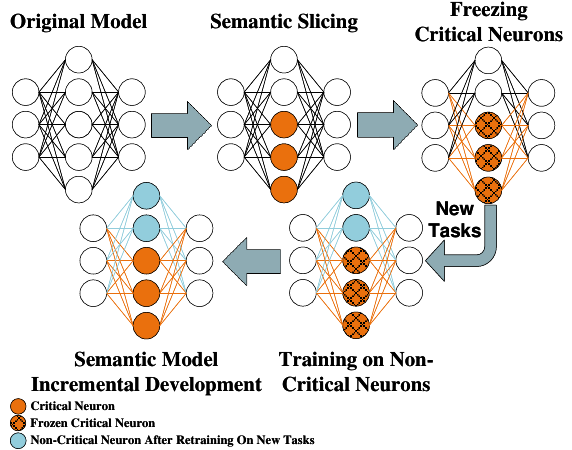}
    \caption{Model Incremental Development Based on Semantic Slicing. We Train Only the Non-Critical Neurons for Model Incremental Development.}
    \label{fig:Evolution}
    \vspace{-18pt}
\end{figure*}

\zsd{
Our strategy enables model incremental development by leveraging the identified critical neurons as stable anchors of previously acquired knowledge. By freezing the weights of these critical neurons during subsequent training phases, the model retains the essential semantic features learned for earlier tasks, thus mitigating catastrophic forgetting. At the same time, the non-critical neurons remain free to adapt, allowing the model to integrate new skills associated with a novel task without diluting its mastery of older tasks.

Concretely, suppose the model has been previously trained on task $A$. We begin by determining the critical neuron set $\mathbf{\Omega}$ that captures task $A$'s essential semantics. These critical neurons, extracted through our earlier semantic slicing procedure, form a robust internal representation of the concepts necessary for solving task $A$. When the model encounters a new task $B$, it relies on these preserved, critical neurons to maintain performance on task $A$ while directing the training process for task $B$ to the non-critical neurons (i.e., those not in $\mathbf{\Omega}$). Figure \ref{fig:Evolution} illustrates this idea: the critical neurons, having proven essential for the original task, remain fixed, while the rest of the network updates to learn the new task.

This approach ensures that the parameters governing the original task’s key semantic representations remain intact, thus safeguarding previously learned knowledge. The model’s representational capacity outside $\mathbf{\Omega}$ is repurposed for incremental learning, enabling it to acquire new skills with minimal interference to existing knowledge. Through this selective updating mechanism, the model incrementally expands its repertoire of competencies in a stable and controlled manner.

Formally, for training on the new task $B$, we adopt the following loss function:
\begin{align}
\label{eq:IncrementalDevelopmentLoss}
\mathcal{L} = \mathcal{L}_{c}(x_B, y_B; \theta),
\end{align}
where $\mathcal{L}_{c}$ denotes the cross-entropy loss and $(x_B, y_B)$ are the training samples and labels from task $B$. Given this objective, the parameter update rule becomes:
\begin{align}
\label{eq:IncrementalDevelopmentParameters}
\theta'_{\mathbf{N}\setminus\mathbf{\Omega}} = \theta_{\mathbf{N}\setminus\mathbf{\Omega}} - \eta \cdot \nabla_{\theta_{\mathbf{N}\setminus\mathbf{\Omega}}}\mathcal{L},
\end{align}
where $\eta$ is the learning rate, $\mathbf{N}$ represents the full set of neurons, and $\mathbf{N}\setminus\mathbf{\Omega}$ is the set of neurons not identified as critical for task $A$.

In this manner, only the parameters not involved in encoding the critical semantics from task $A$ are adjusted, allowing the model to assimilate the new task $B$ without eroding the prior knowledge. During inference, the model employs the frozen, critical parameters for predictions on task $A$, ensuring stable performance on previously learned content. For task $B$, it leverages the updated subset of parameters dedicated to the new skill, allowing the model to switch between tasks while maintaining overall system integrity and performance.}

%% file: Chapters/evaluation.tex
\section{Evaluation}
In this section, we evaluate the performance of \tool{}. We first outline the experimental setup and then we introduce our evaluation aiming to answer the following research questions:
\begin{itemize}[leftmargin=18pt]
\item{\textbf{RQ1:} How does semantic slicing outperform other DNN slicing methods?}
\item{\textbf{RQ2:} Is \tool{} effective for model restructure?}
\item{\textbf{RQ3:} Is \tool{} effective for model re-adaptation?}
\item{\textbf{RQ4:} Is \tool{} effective for model incremental development?}
\end{itemize}

\subsection{Evaluation Setup}
\label{sec:setup}
\zsd{To ensure the comprehensiveness and depth of the experiments, we select four widely utilized datasets in the deep neural network domain—\textbf{MNIST}~\cite{lecun1998gradient}, \textbf{Fashion-MNIST}~\cite{xiao2017fashion}, \textbf{CIFAR-10}~\cite{krizhevsky2009learning}, and \textbf{ImageNet}~\cite{deng2009imagenet}—along with five network models for experimental evaluation: \textbf{MLP}, \textbf{AlexNet}~\cite{krizhevsky2012imagenet}, \textbf{VGG-16}~\cite{simonyan2014very}, \textbf{ResNet-101}~\cite{He_Zhang_Ren_Sun_2016}, and \textbf{VGG-19}~\cite{simonyan2014very}. These models and datasets are not only widely used in the deep learning domain but also extensively applied in software engineering research~\cite{10.1145/3707453, 10.1145/3630011, 10.1145/3672446, 10.1145/3394112, 10.1145/3672454, 10.1145/3644387, 10.1145/3635706}.

\ding{182} {\textbf{MNIST and Fashion-MNIST.} For training, we use a Multilayer Perceptron (MLP) model with two hidden layers containing 512 and 256 neurons, respectively.}

\ding{183} {\textbf{CIFAR-10.} Three deep neural network models—AlexNet, VGG-16, and ResNet-101—are utilized for training on this dataset.}

\ding{184} {\textbf{ImageNet.} Following the setup in~\cite{xie2022npc}, we select data from the top 10 categories for experiments, employing the VGG-19 model.}
}

Based on the datasets and models mentioned, six experimental settings are constructed: MLP-MNIST, MLP-FMNIST, AlexNet-CIFAR10, VGG16-CIFAR10, ResNet101-CIFAR10, and VGG19-ImageNet. These experiments include a variety of configurations, ranging from small to large models and datasets. In each experimental setting, we train at least five different models to repeat the experiments to reduce randomness and ensure robustness. The results are presented as mean values to enhance the experiment reliability and precision.

All these experiments are conducted on the Intel Xeon Silver 4214 Processor with 2 Tesla V100 GPUs with 32GB memory. We have implemented our tool based on Pytorch~\cite{pytorch}. 
The source code, related information of datasets and models, and complete experimental results are released at~\cite{NeuSemSlice2024}.

\subsection{RQ1: The performance of semantic slicing}
In RQ1, we aim to evaluate and compare various slicing methods. Specifically, we follow~\cite{xie2022npc,li2021understanding} to mask the redundant components of the DNN model (\ie, zeroing non-critical neurons) and then compare the model accuracy with only the critical neurons retained. Greater accuracy suggests the slicing method could better identify critical neurons with precision.

\begin{figure*}[h]
    \centering
    \includegraphics[width=0.95\textwidth]{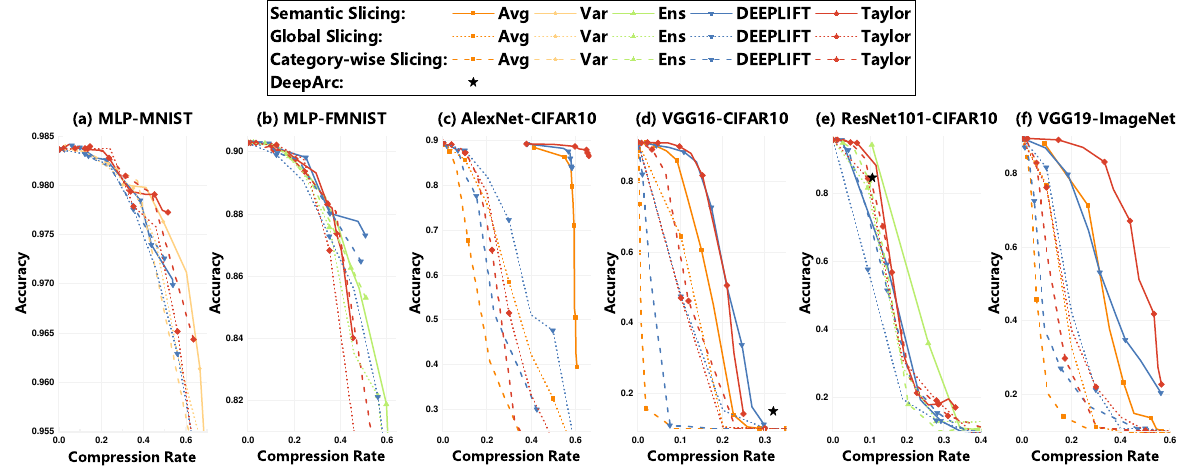}
    \vspace{-8pt}
    \caption{Comparison Results of Performance Across Different Slicing Methods Are Presented. Here, We Only Present the Results for the Top 3 Contribution Metrics for Better Observation.
    Each Dot in the Figure Represents the Model Performance (\ie, Compression Rate and Accuracy) After Masking the Redundant Components of Neurons Identified by a Slicing Method, With DeepArc Represented by a \ding{72}.
    The Lines Representing Semantic Slicing Are Plotted by Varying $\Theta$ Values From 0.9 to 0.99. For Global and Category-Wise Slicing, the Top 10\%, 20\%, Up to 100\% of Neurons per Layer or Category Are Selected for Slicing Respectively. The Results for DeepArc Are Obtained by Removing Redundant Layers With Semantic Similarity Greater Than 0.99. Ideally, Dots (Lines) Located at the \textbf{Upper Right} Are Favored for Their Higher Compression Rates and Better Accuracy Performance, Which Also Suggest That the Corresponding Slicing Methods Better Identify Critical Neurons in the Model.}
    \label{res:rq1}
    \vspace{-18pt}
\end{figure*}

\noindent\textbf{Detailed Configurations.} In RQ1, we compare the following three slicing methods: global slicing, category-wise slicing, and semantic slicing. To assess neuron contribution and select critical neurons, we utilize five neuron contribution metrics: Avg, Var, Ens, DeepLIFT, and Taylor. As outlined in Section \ref{sec:existing_setting}, global slicing and category-wise slicing choose top 10\%, 20\%, ... up to 100\% of neurons in each layer or category for slicing respectively.
Additionally, we introduce DeepArc for comparative analysis. Following the settings used in~\cite{ren2023deeparc}, we set the threshold at 0.99 to identify redundant layers in the model. Subsequently, we remove these redundant layers and compare the model accuracy when only critical semantic layers are retained.

Semantic slicing identifies critical neurons for each category using a set of thresholds \( \Theta = 0.9, 0.91, 0.92, 0.93, 0.94, 0.95, 0.96, 0.97, 0.98, 0.99 \), following the similar settings used by~\cite{ren2023deeparc}. Then it merges them into semantic slices, as described in Section \ref{sec:semslicing}.
For RQ1, the compression rate is defined as the percentage reduction of parameters in the model, while model accuracy denotes the predictive accuracy of the model slices on the test dataset. We conducted comparisons of slicing precision across the previously mentioned six experimental settings.

\noindent\textbf{Comparison with Global Slicing and Category-wise Slicing.} Figure~\ref{res:rq1} compares the slicing precision differences among semantic slicing, global slicing, and category-wise slicing (the complete experimental results can be found in~\cite{NeuSemSlice2024}). For complex models like AlexNet and VGG16, global slicing and category-wise slicing can only maintain certain model performance at very low compression rates, with accuracy significantly declining as the compression rate increases. Semantic slicing demonstrates significant improvement over the other two techniques across all five neuron contribution metrics. For example, in the AlexNet-CIFAR10 case~(Figure~\ref{res:rq1}(c)), semantic slicing uses the neuron contribution metric Taylor to maintain model performance, with only a 0.38\% loss in accuracy at a 50\% compression rate. Conversely, the other two techniques experience significant accuracy losses of around 70\% at the same compression rate.
Similarly, on VGG16 using the DeepLIFT metric~(Figure~\ref{res:rq1}(d)), semantic slicing at around a 10\% compression rate reduces accuracy loss by 41.18\% and 77.77\% compared to global slicing and category-wise slicing.

On larger models like ResNet101 and VGG19~(Figure~\ref{res:rq1}(e) and (f)), global slicing and category-wise slicing are also less effective. Even for datasets like MNIST and Fashion-MNIST~(Figure~\ref{res:rq1}(a) and (b), which require fewer neurons for high accuracy, semantic slicing still slightly outperforms them, maintaining higher model accuracy across all evaluated neuron contribution metrics at the same levels of compression rate.

\begin{table*}[t]
\centering
\caption{Statistical Comparison of Semantic Slicing With Global and Category-Wise Slicing Across Various Experimental Settings.}
\label{res:rq1-1}
\begin{tabularx}{\textwidth}{c >{\centering\arraybackslash}X >{\centering\arraybackslash}X >{\centering\arraybackslash}X >{\centering\arraybackslash}X}
\hline
\multirow{2}{*}{Settings} 
& \multicolumn{2}{c}{Semantic vs Global} 
& \multicolumn{2}{c}{Semantic vs Category-wise} \\ \cline{2-5} 
& $p$-value & Cohen's $d$ & $p$-value & Cohen's $d$ \\ \hline
MLP-MNIST & $**$ & $**$ & $*$ & $*$ \\
MLP-FMNIST & $***$ & $***$ & $**$ & $**$ \\
AlexNet-CIFAR10 & $***$ & $***$ & $***$ & $***$ \\
VGG16-CIFAR10 & $**$ & $***$ & $***$ & $***$ \\
ResNet101-CIFAR10 & $***$ & $***$ & $***$ & $***$ \\
VGG19-ImageNet & $***$ & $***$ & $***$ & $***$ \\ \hline
\end{tabularx}
\begin{tablenotes}
\footnotesize
\item $p$-value: *** < 0.01 (highly significant), ** < 0.05 (significant), * > 0.05 (not significant).
\item Cohen's $d$: *** (> 0.8, large effect), ** (0.5 - 0.8, medium effect), * (0.2 - 0.5, small effect).
\end{tablenotes}
\end{table*}

\zsd{To further clarify the extent of the differences among these approaches, we conduct paired \(t\)-tests and compute effect sizes using Cohen’s $d$ on the AUC values obtained within the same range from the compression rate-accuracy curves (see Table~\ref{res:rq1-1}). These tests confirm that semantic slicing achieves statistically significant improvements over global slicing in all experimental configurations ($p$-values < 0.05) and, except for the MLP-MNIST setting, also significantly outperforms category-wise slicing. Moreover, the corresponding effect sizes are medium to large in most cases, indicating that the advantages of semantic slicing are not only statistically robust but also practically meaningful.}

These results adequately confirm that an inadequate account of the uneven distribution of critical neurons across different categories and network layers leads to poor slicing performance. This highlights the advantage of the proposed technique, which can dynamically identify critical features and components of the model based on semantic similarity.

\noindent\textbf{The Effects of DeepArc.} For DeepArc, we identify redundant layers in six settings based on layer similarity, as shown in Figure \ref{res:rq1}. Firstly, for the MLP, the model only has two hidden layers. When any layer is removed, we follow the model restructuring detailed in DeepArc and introduce a ``sewing layer'' to align the input and output of the layers. However, this results in no change in the size of the model, indicating that DeepArc cannot effectively identify redundancies in such models. In the other four configurations, DeepArc only identifies redundant layers with a similarity greater than 0.99 in VGG16 and ResNet101. 

For AlexNet, the maximum layer similarity is 0.95. If the threshold is lowered to 0.95 and the corresponding redundant layer in the module is removed, a fully connected layer containing 87.97\% of the model's parameters is eliminated, reducing the model accuracy to 10.94\%. The results for VGG19 are similar, with the highest layer similarity being only 0.97. For VGG16, removing redundant layers within the module leads to a model compression rate of 32.06\% but an accuracy of only 14.84\%. For ResNet101, after removing 10.68\% of redundant layers, the accuracy remains at 84.77\%. However, its precision in identifying critical model components is still not as high as that of \tool{}. These results suggest that DeepArc is relatively effective only in large models with small datasets, due to the difficulty in sharing semantics in smaller models or models with lower redundancy.

\noindent\textbf{Comparing different Neuron Contribution Metrics.} Our \tool{} framework is orthogonal to the neuron contribution metrics, meaning it can be integrated seamlessly with a range of contribution metrics. Through our framework, we can fairly compare different neuron contribution metrics. We can observe that in relatively simple models like MLP, the performance of these metrics is close. They all effectively assess neuron contributions, enabling the precise selection of critical neurons. For example, in the case of MLP-MNIST, even at a compression rate of 70\%, the model's accuracy only drops by approximately 4\%. In contrast, for more complex models such as AlexNet and VGG16, DeepLIFT and Taylor metrics perform better. For AlexNet, the Taylor metric achieves a mere 3\% accuracy drop at a compression rate of 66.05\%; for VGG16, both DeepLIFT and Taylor metrics maintain the model's accuracy within a 2\% decrease at a compression rate of approximately 10\%. For larger models like ResNet101 and VGG19, the performance of the metrics is similarly close in ResNet101, with Ens achieving only a 1.54\% drop in accuracy at a 10.56\% compression rate, while in VGG19, Taylor significantly outperforms other metrics, maintaining a 2.54\% accuracy drop at a 24.67\% compression rate. \looseness=-1

The Var metric exhibits lower accuracy after model compression, which is especially evident in AlexNet and VGG16. We also calculate the time required to compute various metrics: Avg, Var, Ens, DeepLIFT, and Taylor, which take 5.4s, 5.5s, 5.6s, 30.9s, and 6.2s respectively on the MNIST dataset. In comparison to other methods, DeepLIFT exhibits significantly higher time consumption.
\textit{Overall, the DeepLIFT and Taylor metrics outperform the other three metrics, while DeepLIFT is less efficient.}

\vspace{-0.2cm}
\begin{tcolorbox}[size=title]{
\textbf{Answer to RQ1:  Compared to global slicing and category-wise slicing, semantic slicing better identifies critical neurons with precision. At 50\% compression rate, semantic slicing increases model performance from 23.24\% and 11.55\% to 89.20\% compared to two other techniques respectively (see AlexNet-CIFAR10, neuron contribution metric Taylor, Figure~\ref{res:rq1}(c)).}}
\end{tcolorbox}

\vspace{-0.3cm}
\subsection{RQ2: Model Restructure}
\begin{table*}[]
\centering
\caption{\zsd{Results of Model Restructure.}}
\vspace{-10pt}
\label{rq2}
\resizebox{\textwidth}{!}{%
\begin{tabular}{ccccccccccccccccccc}
\hline
 & \multicolumn{3}{c}{MLP-MNIST} & \multicolumn{3}{c}{MLP-FMNIST} & \multicolumn{3}{c}{AlexNet-CIFAR10} & \multicolumn{3}{c}{VGG16-CIFAR10} & \multicolumn{3}{c}{ResNet101-CIFAR10} & \multicolumn{3}{c}{VGG19-ImageNet} \\ \hline
\multicolumn{1}{c|}{\begin{tabular}[c]{@{}c@{}}Compression\\ Algorithm\end{tabular}} & CR \textcolor{red}{$\uparrow$} & NI \textcolor{green}{$\downarrow$} & \multicolumn{1}{c|}{MA \textcolor{red}{$\uparrow$}} & CR \textcolor{red}{$\uparrow$}  & NI \textcolor{green}{$\downarrow$} & \multicolumn{1}{c|}{MA \textcolor{red}{$\uparrow$}} & CR \textcolor{red}{$\uparrow$} & NI \textcolor{green}{$\downarrow$} & \multicolumn{1}{c|}{MA \textcolor{red}{$\uparrow$}} & CR \textcolor{red}{$\uparrow$} & NI \textcolor{green}{$\downarrow$} & \multicolumn{1}{c|}{MA \textcolor{red}{$\uparrow$}} & CR \textcolor{red}{$\uparrow$} & NI \textcolor{green}{$\downarrow$} & \multicolumn{1}{c|}{MA \textcolor{red}{$\uparrow$}} & CR \textcolor{red}{$\uparrow$} & NI \textcolor{green}{$\downarrow$} & MA \textcolor{red}{$\uparrow$} \\ \hline

\multicolumn{1}{c|}{LAMP} & 52.17\% & 7 & \multicolumn{1}{c|}{98.26\%} & 36.97\% & 9 & \multicolumn{1}{c|}{90.01\%} & 30.17\% & 7 & \multicolumn{1}{c|}{89.49\%} & 18.55\% & 4 & \multicolumn{1}{c|}{91.28\%} & 45.96\% & 12 & \multicolumn{1}{c|}{90.77\%} & 30.17\% & 7 & \textbf{89.54\%} \\

\multicolumn{1}{c|}{Global+LAMP} & 48.27\% & 5 & \multicolumn{1}{c|}{98.26\%} & 38.82\% & 7 & \multicolumn{1}{c|}{89.78\%} & 33.85\% & 6 & \multicolumn{1}{c|}{87.87\%} & 14.65\% & 1 & \multicolumn{1}{c|}{46.84\%} & 45.79\% & 10 & \multicolumn{1}{c|}{76.29\%} & 33.86\% & 6 & 76.31\% \\

\multicolumn{1}{c|}{Category-wise+LAMP} & 50.82\% & 6 & \multicolumn{1}{c|}{98.30\%} & 39.19\% & 8 & \multicolumn{1}{c|}{90.04\%} & 33.95\% & 6 & \multicolumn{1}{c|}{82.12\%} & 20.08\% & 4 & \multicolumn{1}{c|}{90.31\%} & 46.03\% & 10 & \multicolumn{1}{c|}{81.27\%} & 33.92\% & 6 & 58.42\% \\

\multicolumn{1}{c|}{\tool{}+LAMP} & \textbf{53.05\%} & 5 & \multicolumn{1}{c|}{\textbf{98.32\%}} & \textbf{39.70\%} & 8 & \multicolumn{1}{c|}{\textbf{90.21\%}} & \textbf{34.88\%} & 6 & \multicolumn{1}{c|}{\textbf{89.52\%}} & \textbf{20.36\%} & 4 & \multicolumn{1}{c|}{\textbf{91.29\%}} & \textbf{46.46\%} & 10 & \multicolumn{1}{c|}{\textbf{92.11\%}} & \textbf{37.25\%} & 6 & 89.19\% \\ \hline

\multicolumn{1}{c|}{Glob} & 40.95\% & 5 & \multicolumn{1}{c|}{98.31\%} & 33.66\% & 8 & \multicolumn{1}{c|}{90.09\%} & 60.28\% & 18 & \multicolumn{1}{c|}{\textbf{89.44\%}} & 22.62\% & 5 & \multicolumn{1}{c|}{91.25\%} & 53.67\% & 15 & \multicolumn{1}{c|}{72.21\%} & 22.62\% & 5 & \textbf{89.42\%} \\

\multicolumn{1}{c|}{Global+Glob} & 36.13\% & 3 & \multicolumn{1}{c|}{98.33\%} & 35.60\% & 6 & \multicolumn{1}{c|}{89.80\%} & 62.37\% & 17 & \multicolumn{1}{c|}{87.75\%} & 14.65\% & 1 & \multicolumn{1}{c|}{47.00\%} & 53.53\% & 13 & \multicolumn{1}{c|}{46.29\%} & 22.86\% & 3 & 76.15\% \\

\multicolumn{1}{c|}{Category-wise+Glob} & 39.28\% & 4 & \multicolumn{1}{c|}{98.31\%} & 32.62\% & 6 & \multicolumn{1}{c|}{90.18\%} & 62.43\% & 17 & \multicolumn{1}{c|}{81.99\%} & 24.07\% & 5 & \multicolumn{1}{c|}{90.28\%} & 53.73\% & 13 & \multicolumn{1}{c|}{47.82\%} & 22.93\% & 3 & 58.15\% \\

\multicolumn{1}{c|}{\tool{}+Glob} & \textbf{42.04\%} & 3 & \multicolumn{1}{c|}{\textbf{98.37\%}} & \textbf{36.52\%} & 7 & \multicolumn{1}{c|}{\textbf{90.18\%}} & \textbf{62.96\%} & 17 & \multicolumn{1}{c|}{\textbf{89.44\%}} & \textbf{24.34\%} & 5 & \multicolumn{1}{c|}{\textbf{91.28\%}} & \textbf{54.10\%} & 13 & \multicolumn{1}{c|}{\textbf{73.96\%}} & \textbf{26.93\%} & 3 & 89.12\% \\ \hline

\multicolumn{1}{c|}{Unif} & 40.95\% & 5 & \multicolumn{1}{c|}{98.34\%} & 40.13\% & 10 & \multicolumn{1}{c|}{89.81\%} & 55.99\% & 16 & \multicolumn{1}{c|}{88.59\%} & 14.26\% & 3 & \multicolumn{1}{c|}{91.24\%} & 51.23\% & 14 & \multicolumn{1}{c|}{83.95\%} & 55.99\% & 16 & 85.85\% \\

\multicolumn{1}{c|}{Global+Unif} & 36.13\% & 3 & \multicolumn{1}{c|}{98.29\%} & 41.88\% & 8 & \multicolumn{1}{c|}{89.64\%} & 58.31\% & 15 & \multicolumn{1}{c|}{87.08\%} & 14.65\% & 1 & \multicolumn{1}{c|}{47.19\%} & 51.08\% & 12 & \multicolumn{1}{c|}{55.52\%} & 53.81\% & 13 & 67.19\% \\

\multicolumn{1}{c|}{Category-wise+Unif} & 39.28\% & 4 & \multicolumn{1}{c|}{98.28\%} & 39.19\% & 8 & \multicolumn{1}{c|}{89.89\%} & 58.37\% & 15 & \multicolumn{1}{c|}{81.60\%} & 15.87\% & 3 & \multicolumn{1}{c|}{90.21\%} & 51.29\% & 12 & \multicolumn{1}{c|}{58.68\%} & 53.86\% & 13 & 46.46\% \\

\multicolumn{1}{c|}{\tool{}+Unif} & \textbf{42.04\%} & 3 & \multicolumn{1}{c|}{\textbf{98.36\%}} & \textbf{42.71\%} & 9 & \multicolumn{1}{c|}{\textbf{89.95\%}} & \textbf{58.96\%} & 15 & \multicolumn{1}{c|}{\textbf{88.68\%}} & \textbf{16.17\%} & 3 & \multicolumn{1}{c|}{\textbf{91.31\%}} & \textbf{51.68\%} & 12 & \multicolumn{1}{c|}{\textbf{84.03\%}} & \textbf{56.25\%} & 13 & \textbf{87.50\%} \\ \hline

\multicolumn{1}{c|}{ERK} & 34.39\% & 4 & \multicolumn{1}{c|}{98.30\%} & 30.17\% & 7 & \multicolumn{1}{c|}{90.08\%} & 55.99\% & 16 & \multicolumn{1}{c|}{89.37\%} & 26.49\% & 6 & \multicolumn{1}{c|}{91.19\%} & 14.26\% & 3 & \multicolumn{1}{c|}{\textbf{95.66\%}} & 26.49\% & 6 & \textbf{89.58\%} \\

\multicolumn{1}{c|}{Global+ERK} & 36.13\% & 3 & \multicolumn{1}{c|}{98.31\%} & 32.21\% & 5 & \multicolumn{1}{c|}{89.73\%} & 58.31\% & 15 & \multicolumn{1}{c|}{87.66\%} & 14.65\% & 1 & \multicolumn{1}{c|}{46.95\%} & 22.38\% & 3 & \multicolumn{1}{c|}{80.87\%} & 30.38\% & 5 & 76.23\% \\

\multicolumn{1}{c|}{Category-wise+ERK} & 32.53\% & 3 & \multicolumn{1}{c|}{98.34\%} & 32.62\% & 6 & \multicolumn{1}{c|}{90.18\%} & 58.37\% & 15 & \multicolumn{1}{c|}{82.15\%} & 27.87\% & 6 & \multicolumn{1}{c|}{90.34\%} & 22.72\% & 3 & \multicolumn{1}{c|}{82.84\%} & 30.45\% & 5 & 58.31\% \\

\multicolumn{1}{c|}{\tool{}+ERK} & \textbf{42.04\%} & 3 & \multicolumn{1}{c|}{\textbf{98.33\%}} & \textbf{33.18\%} & 6 & \multicolumn{1}{c|}{\textbf{90.24\%}} & \textbf{58.96\%} & 15 & \multicolumn{1}{c|}{\textbf{89.41\%}} & \textbf{28.13\%} & 6 & \multicolumn{1}{c|}{\textbf{91.28\%}} & \textbf{23.34\%} & 3 & \multicolumn{1}{c|}{93.97\%} & \textbf{34.06\%} & 5 & 89.19\% \\ \hline
\end{tabular}
}
\end{table*}

\noindent\textbf{Detailed Configuration.} To systematically assess \tool{} for the model restructure task, we adopt the same experimental setup detailed in~\cite{ren2023deeparc}, and validate by comparing the compression efficiency and effects on the model before and after applying \tool{}. Specifically, after removing the non-critical neurons of the model, we expect the compression technology to directly focus on the redundant parts within the critical neurons, improving the iteration efficiency\footnote{It is important to note that in this experiment and the subsequent RQ3 and RQ4, our method is based on the optimal configuration identified in RQ1, while global slicing and category-wise slicing use configurations with similar slice sizes under the same neuron contribution metric.}. 
Based on this, we apply four of the most advanced model compression techniques currently available—LAMP~\cite{Lee_Park_Mo_Ahn_Shin_2020}, Global~\cite{Morcos_Yu_Paganini_Tian_2019}, Uniform+~\cite{Gale_Elsen_Hooker_2019}, and ERK~\cite{Evci_Gale_Menick_Castro_Elsen_2019}—for optimal compression results. \zsd{Additionally, we also present the results of global slicing and category-wise slicing for comparison.}

In our experiments, we take the following metrics to evaluate the performance of model compression. The Compression Rate (\textbf{CR}) indicates the reduction in parameter size, the Number of Iterations (\textbf{NI}) reflects the efficiency of the compression process, and Model Accuracy (\textbf{MA}) measures the predictive performance on the test dataset. 

\noindent\textbf{Results and Findings.} Table \ref{rq2} presents the experimental results of \tool{} in model restructure. Across all experimental settings, the application of \tool{} in model restructure allows models to achieve higher compression ratios with fewer iterations, indicating that \tool{} can improve both compression efficiency and effectiveness. Apart from the VGG19-ImageNet configuration, \tool{} not only enhances compression efficiency and rates but also slightly improves model accuracy. Specifically, in six experimental settings, the average number of iterations required for model compression decreases from 8.67 to 7.50, enhancing efficiency by 13.46\%, and the model compression ratio increases from 37.26\% to 40.26\%. 

\zsd{For global slicing and category-wise slicing, we observe that in simpler models like MLP-MNIST and MLP-FMNIST, their performance is similar to \tool{}, both effectively improving compression efficiency. However, as the model complexity increases, \tool{}'s ability to accurately identify critical neurons becomes more apparent. In the case of AlexNet-CIFAR10, with the same number of compression iterations and similar compression ratios, \tool{} achieves an average improvement of 1.67\% in model accuracy compared to global slicing and a 7.30\% improvement compared to category-wise slicing. For more complex models, such as VGG16-CIFAR10, ResNet101-CIFAR10, and VGG19-ImageNet, global slicing and category-wise slicing often result in significant accuracy drops. For instance, global slicing reduces the accuracy of VGG16-CIFAR10 to around 47\%, while category-wise slicing drops the accuracy of VGG19-ImageNet to about 58\%, whereas \tool{} does not experience such issues.} These results demonstrate that the application of \tool{} in model restructure boosts the performance of model compression and enhances the performance of compressed models.

\begin{tcolorbox}[size=title]{
\textbf{Answer to RQ2: } \textbf{For model restructure, \tool{} not only elevates the efficiency of compression technology—an average increase of 13.46\%—but also achieves approximately a 3\% enhancement in the compression rate and a slight increase in accuracy.}}
\end{tcolorbox}

\subsection{RQ3: Model Re-adaptation}
\label{sec:re-adaptation}

\begin{table*}[]
\centering
\caption{\zsd{Results of Model Re-adaptation.}}
\vspace{-10pt}
\label{table:rq3}
\resizebox{\textwidth}{!}{%
\begin{tabular}{cccccccccccccccccccc}
\hline
 &  & \multicolumn{3}{c}{MLP-MNIST} & \multicolumn{3}{c}{MLP-FMNIST} & \multicolumn{3}{c}{AlexNet-CIFAR10} & \multicolumn{3}{c}{VGG16-CIFAR10} & \multicolumn{3}{c}{ResNet101-CIFAR10} & \multicolumn{3}{c}{VGG19-ImageNet} \\ \hline
 & \multicolumn{1}{c|}{Method} & TPR \textcolor{green}{$\downarrow$} & RA \textcolor{red}{$\uparrow$} & \multicolumn{1}{c|}{RAA} \textcolor{red}{$\uparrow$} & TPR \textcolor{green}{$\downarrow$} & RA \textcolor{red}{$\uparrow$} & RAA \textcolor{red}{$\uparrow$} & TPR \textcolor{green}{$\downarrow$} & RA \textcolor{red}{$\uparrow$} & RAA \textcolor{red}{$\uparrow$} & TPR \textcolor{green}{$\downarrow$} & RA \textcolor{red}{$\uparrow$} & RAA \textcolor{red}{$\uparrow$} & TPR \textcolor{green}{$\downarrow$} & RA \textcolor{red}{$\uparrow$} & RAA \textcolor{red}{$\uparrow$} & TPR \textcolor{green}{$\downarrow$} & RA \textcolor{red}{$\uparrow$} & RAA \textcolor{red}{$\uparrow$} \\ \hline
\multirow{4}{*}{FGSM} & \multicolumn{1}{c|}{Fully Training} & 100.00\% & 100.00\% & \multicolumn{1}{c|}{14.20\%} & 100.00\% & \textbf{98.25\%} & \multicolumn{1}{c|}{10.10\%} & 100.00\% & 98.40\% & \multicolumn{1}{c|}{43.90\%} & 100.00\% & 97.83\% & \multicolumn{1}{c|}{13.10\%} & 100.00\% & 100.00\% & \multicolumn{1}{c|}{49.50\%} & 100.00\% & 98.54\% & \textbf{48.59\%} \\

 & \multicolumn{1}{c|}{DeepArc} & 75.01\% & 100.00\% & \multicolumn{1}{c|}{13.40\%} & 75.01\% & 97.18\% & \multicolumn{1}{c|}{11.20\%} & 99.18\% & 99.43\% & \multicolumn{1}{c|}{50.40\%} & 67.60\% & 98.80\% & \multicolumn{1}{c|}{13.70\%} & 56.26\% & 100.00\% & \multicolumn{1}{c|}{53.60\%} & 95.82\% & 98.44\% & 42.66\% \\

 & \multicolumn{1}{c|}{Global Slicing} & 33.98\% & 100.00\% & \multicolumn{1}{c|}{13.60\%} & 54.06\% & 95.00\% & \multicolumn{1}{c|}{11.30\%} & 40.10\% & 98.60\% & \multicolumn{1}{c|}{41.10\%} & 89.85\% & 97.05\% & \multicolumn{1}{c|}{16.20\%} & 90.56\% & 90.40\% & \multicolumn{1}{c|}{52.60\%} & 89.97\% & 98.44\% & \textbf{48.59\%} \\

  & \multicolumn{1}{c|}{Category-wise Slicing} & 37.57\% & 100.00\% & \multicolumn{1}{c|}{15.90\%} & 51.14\% & 95.85\% & \multicolumn{1}{c|}{11.10\%} & 46.65\% & 98.35\% & \multicolumn{1}{c|}{45.30\%} & 88.05\% & 95.20\% & \multicolumn{1}{c|}{13.60\%} & 87.22\% & 91.40\% & \multicolumn{1}{c|}{49.30\%} & 89.89\% & 95.47\% & 46.88\% \\
 
 & \multicolumn{1}{c|}{\tool{}} & 26.74\% & 100.00\% & \multicolumn{1}{c|}{18.10\%} & 49.26\% & 97.72\% & \multicolumn{1}{c|}{11.40\%} & 33.95\% & 99.55\% & \multicolumn{1}{c|}{46.10\%} & 87.41\% & 99.60\% & \multicolumn{1}{c|}{15.70\%} & 86.35\% & 100.00\% & \multicolumn{1}{c|}{54.20\%} & 85.22\% & 98.78\% & 47.66\% \\
 
 & \multicolumn{1}{c|}{\tool{}+DeepArc} & \textbf{24.17\%} & 100.00\% & \multicolumn{1}{c|}{\textbf{21.20\%}} & \textbf{40.29\%} & 98.00\% & \multicolumn{1}{c|}{\textbf{11.50\%}} & \textbf{33.21\%} & \textbf{100.00\%} & \multicolumn{1}{c|}{\textbf{52.70\%}} & \textbf{60.39\%} & \textbf{99.88\%} & \multicolumn{1}{c|}{\textbf{37.30\%}} & \textbf{47.30\%} & 100.00\% & \multicolumn{1}{c|}{\textbf{54.60\%}} & \textbf{81.12\%} & \textbf{98.85\%} & 44.22\% \\ \hline
\multirow{4}{*}{PGD} & \multicolumn{1}{c|}{Fully Training} & 100.00\% & 100.00\% & \multicolumn{1}{c|}{\textbf{14.00\%}} & 100.00\% & 97.90\% & \multicolumn{1}{c|}{6.00\%} & 100.00\% & 97.68\% & \multicolumn{1}{c|}{1.30\%} & 100.00\% & 95.80\% & \multicolumn{1}{c|}{\textbf{2.90\%}} & 100.00\% & 99.30\% & \multicolumn{1}{c|}{0.90\%} & 100.00\% & 99.38\% & 22.81\% \\

 & \multicolumn{1}{c|}{DeepArc} & 75.01\% & 100.00\% & \multicolumn{1}{c|}{12.00\%} & 75.01\% & 97.37\% & \multicolumn{1}{c|}{7.20\%} & 99.18\% & \textbf{99.95\%} & \multicolumn{1}{c|}{0.10\%} & 67.60\% & 99.53\% & \multicolumn{1}{c|}{0.00\%} & 56.26\% & 95.30\% & \multicolumn{1}{c|}{4.20\%} & 95.82\% & 99.11\% & 20.78\% \\

 & \multicolumn{1}{c|}{Global Slicing} & 33.98\% & 100.00\% & \multicolumn{1}{c|}{11.90\%} & 54.06\% & 95.95\% & \multicolumn{1}{c|}{8.60\%} & 40.10\% & 96.70\% & \multicolumn{1}{c|}{1.00\%} & 89.85\% & 94.85\% & \multicolumn{1}{c|}{0.00\%} & 90.56\% & 92.90\% & \multicolumn{1}{c|}{2.60\%} & 89.97\% & 96.41\% & 17.50\% \\

  & \multicolumn{1}{c|}{Category-wise Slicing} & 37.57\% & 100.00\% & \multicolumn{1}{c|}{11.30\%} & 51.14\% & 95.65\% & \multicolumn{1}{c|}{\textbf{10.50\%}} & 46.65\% & 96.20\% & \multicolumn{1}{c|}{0.90\%} & 88.05\% & 96.40\% & \multicolumn{1}{c|}{0.00\%} & 87.22\% & 96.40\% & \multicolumn{1}{c|}{3.10\%} & 89.89\% & 95.63\% & 17.34\% \\
  
 & \multicolumn{1}{c|}{\tool{}} & 26.74\% & 100.00\% & \multicolumn{1}{c|}{12.90\%} & 49.26\% & 97.90\% & \multicolumn{1}{c|}{5.80\%} & 33.95\% & 99.30\% & \multicolumn{1}{c|}{\textbf{1.40\%}} & 87.41\% & 99.75\% & \multicolumn{1}{c|}{0.00\%} & 86.35\% & \textbf{99.40\%} & \multicolumn{1}{c|}{1.40\%} & 85.22\% & 99.18\% & 20.16\% \\
 
 & \multicolumn{1}{c|}{\tool{}+DeepArc} & \textbf{24.17\%} & 100.00\% & \multicolumn{1}{c|}{12.60\%} & \textbf{40.29\%} & \textbf{97.92\%} & \multicolumn{1}{c|}{8.10\%} & \textbf{33.21\%} & 98.98\% & \multicolumn{1}{c|}{0.30\%} & \textbf{60.39\%} & \textbf{99.83\%} & \multicolumn{1}{c|}{0.00\%} & \textbf{47.30\%} & 98.93\% & \multicolumn{1}{c|}{\textbf{4.40\%}} & \textbf{81.12\%} & \textbf{99.41\%} & \textbf{24.06\%} \\ \hline
\end{tabular}
}
\vspace{-0.1cm}
\end{table*}
\noindent\textbf{Detailed Configuration.}
To answer RQ3, we conduct an experiment using two prevalently-employed adversarial sample generation techniques, Fast Gradient Sign Method~(FGSM)~\cite{goodfellow2014explaining} and Projected Gradient Descent~(PGD)~\cite{madry2017towards}. We generate 1,000 adversarial samples, with FGSM reducing the model's average accuracy to 18.53\%, and PGD to 3.8\%. 
We aim to compare the Training Parameter Ratio~(\textbf{TPR}) and the Retrained Accuracy~(\textbf{RA})  of existing approaches and our proposed method. Specifically, the experiment settings include fully training~(i.e., comparison baseline), DeepArc, \zsd{global slicing, category-wise slicing}, \tool{}, and a combination of \tool{} with DeepArc\footnote{The term ``a combination of \tool{} with DeepArc'' refers to retraining the critical neurons identified by our method in the critical modules recognized by DeepArc.} under identical training epochs. Additionally, this RQ further assesses the resilience of each method against similar adversarial attacks. We generate another 1,000 adversarial samples after the model re-adaptation, and obtain the model accuracy against the samples (\ie, shown as Re-Attack Accuracy, \textbf{RAA}).

\noindent\textbf{Results and Findings.} Table \ref{table:rq3} presents our experimental results in the six settings. Compared to the traditional method of fully training and DeepArc, our method demonstrates a significant reduction in the required model parameters for training, averaging only 61.49\%. This represents an average decrease of 38.51\% in training parameters compared to the fully training strategy and an average reduction of 16.66\% relative to DeepArc. Moreover, our method surpasses fully training by an average of 0.67\% and DeepArc by an average of 0.50\% in Retrained Accuracy, demonstrating an advantage in both training cost-efficiency and effectiveness. \zsd{Compared to global slicing and category-wise slicing, with similar training parameters, our method shows an average improvement of 2.91\% and 2.89\% in Retrained Accuracy, respectively. This highlights that the inaccurate identification of critical neurons leads to reduced Re-adaptation effectiveness. Although fewer training parameters are used than the fully training method, the training accuracy also decreases accordingly. }Among the six methods, our method combined with DeepArc achieves a 0.73\% increase in Retrained Accuracy compared to traditional training methods, with the lowest training cost~(averaging only 47.75\% of the parameters trained).  

Our method combined with DeepArc also shows the best performance in terms of re-attack performance, achieving up to a 17.8\% improvement in RAA. The results further demonstrate the superior performance of our method in model re-adaptation scenarios. However, we observe that all of the methods experience notable performance loss against re-attack, alerting us to design better methods against adversarial attacks in the future, to be discussed in Section~\ref{sec:threats-validity}. Finally, \tool{} is comparable to DeepArc in efficiency, requiring only 93.53\% of the time needed for full training on average. When \tool{} is combined with DeepArc, it requires only 89.80\% of the full training time. The detailed data is available at~\cite{NeuSemSlice2024}.
 
\begin{tcolorbox}[size=title]{\textbf{Answer to RQ3: Our approach reduces the cost of model re-adaptation by maintaining critical neurons, requiring only an average of 61.49\% of the parameters. Our method also exhibits slight advantages in terms of efficiency and model accuracy over baselines.}}
\end{tcolorbox}

\subsection{RQ4: Model Incremental Development}
\label{sec:incremental-develop}

\begin{table*}[]
\centering
\caption{\zsd{Results of Model Incremental Development(Scenario I)}}
\vspace{-10pt}
\label{rq4}
\resizebox{\textwidth}{!}{
\begin{tabular}{ccccccccccccccccccc}
\hline
 & \multicolumn{3}{c}{MLP-MNIST} & \multicolumn{3}{c}{MLP-FMNIST} & \multicolumn{3}{c}{AlexNet-CIFAR10} & \multicolumn{3}{c}{VGG16-CIFAR10} & \multicolumn{3}{c}{ResNet101-CIFAR10} & \multicolumn{3}{c}{VGG19-ImageNet} \\ \hline
\multicolumn{1}{c|}{Method} & TaskA \textcolor{red}{$\uparrow$} & TaskB \textcolor{red}{$\uparrow$} & \multicolumn{1}{c|}{Avg \textcolor{red}{$\uparrow$}} & TaskA \textcolor{red}{$\uparrow$} & TaskB \textcolor{red}{$\uparrow$} & \multicolumn{1}{c|}{Avg \textcolor{red}{$\uparrow$}} & TaskA \textcolor{red}{$\uparrow$} & TaskB \textcolor{red}{$\uparrow$} & \multicolumn{1}{c|}{Avg \textcolor{red}{$\uparrow$}} & TaskA \textcolor{red}{$\uparrow$} & TaskB \textcolor{red}{$\uparrow$} & \multicolumn{1}{c|}{Avg \textcolor{red}{$\uparrow$}} & TaskA \textcolor{red}{$\uparrow$} & TaskB \textcolor{red}{$\uparrow$} & \multicolumn{1}{c|}{Avg \textcolor{red}{$\uparrow$}} & TaskA \textcolor{red}{$\uparrow$} & TaskB \textcolor{red}{$\uparrow$} & Avg \textcolor{red}{$\uparrow$} \\ \hline
\multicolumn{1}{c|}{Original} & 98.92\% & 0.00\% & \multicolumn{1}{c|}{49.46\%} & 97.44\% & 0.00\% & \multicolumn{1}{c|}{48.72\%} & 89.14\% & 0.00\% & \multicolumn{1}{c|}{44.57\%} & 92.52\% & 0.00\% & \multicolumn{1}{c|}{46.26\%} & 96.90\% & 0.00\% & \multicolumn{1}{c|}{48.45\%} & 92.15\% & 0.00\% & 46.08\% \\

\multicolumn{1}{c|}{Retrain} & 0.00\% & 98.50\% & \multicolumn{1}{c|}{49.25\%} & 2.78\% & 83.28\% & \multicolumn{1}{c|}{43.03\%} & 0.00\% & 89.40\% & \multicolumn{1}{c|}{44.70\%} & 0.00\% & 90.80\% & \multicolumn{1}{c|}{45.40\%} & 0.00\% & 85.86\% & \multicolumn{1}{c|}{42.93\%} & 5.37\% & 88.92\% & 47.15\% \\

\multicolumn{1}{c|}{Replay 5\%} & 87.54\% & 98.32\% & \multicolumn{1}{c|}{92.93\%} & 83.96\% & 83.14\% & \multicolumn{1}{c|}{83.55\%} & 46.32\% & 88.94\% & \multicolumn{1}{c|}{67.63\%} & 54.42\% & 90.76\% & \multicolumn{1}{c|}{72.59\%} & 26.54\% & 85.00\% & \multicolumn{1}{c|}{55.77\%} & 85.62\% & 87.77\% & 86.69\% \\

\multicolumn{1}{c|}{Replay 10\%} & 91.86\% & 98.13\% & \multicolumn{1}{c|}{95.00\%} & 87.90\% & 83.20\% & \multicolumn{1}{c|}{85.55\%} & 59.70\% & 88.16\% & \multicolumn{1}{c|}{73.93\%} & 62.80\% & 90.12\% & \multicolumn{1}{c|}{76.46\%} & 39.40\% & 84.68\% & \multicolumn{1}{c|}{62.04\%} & 87.23\% & 88.38\% & 87.81\% \\

\multicolumn{1}{c|}{Replay 100\%} & 97.20\% & 97.48\% & \multicolumn{1}{c|}{97.34\%} & 94.44\% & 81.78\% & \multicolumn{1}{c|}{88.11\%} & 82.96\% & 85.66\% & \multicolumn{1}{c|}{84.31\%} & 86.98\% & 88.30\% & \multicolumn{1}{c|}{87.64\%} & 84.12\% & 90.18\% & \multicolumn{1}{c|}{87.15\%} & 90.46\% & 89.38\% & 89.92\% \\

\multicolumn{1}{c|}{DeepArc} & / & / & \multicolumn{1}{c|}{/} & / & / & \multicolumn{1}{c|}{/} & 0.00\% & 84.84\% & \multicolumn{1}{c|}{42.42\%} & 20.44\% & 0.00\% & \multicolumn{1}{c|}{10.22\%} & 46.68\% & 0.00\% & \multicolumn{1}{c|}{23.34\%} & 0.00\% & 84.92\% & 42.46\% \\

\multicolumn{1}{c|}{Global Slicing} & 93.58\% & 97.75\% & \multicolumn{1}{c|}{95.66\%} & 82.10\% & 78.46\% & \multicolumn{1}{c|}{80.28\%} & 69.88\% & 87.26\% & \multicolumn{1}{c|}{78.57\%} & 65.74\% & 86.44\% & \multicolumn{1}{c|}{76.09\%} & 96.92\% & 79.20\% & \multicolumn{1}{c|}{88.06\%} & 88.38\% & 89.46\% & 88.92\% \\

\multicolumn{1}{c|}{Category-wise Slicing} & 96.85\% & 97.44\% & \multicolumn{1}{c|}{97.14\%} & 82.12\% & 81.60\% & \multicolumn{1}{c|}{81.86\%} & 69.68\% & 87.18\% & \multicolumn{1}{c|}{78.43\%} & 72.96\% & 86.54\% & \multicolumn{1}{c|}{79.75\%} & 95.34\% & 78.98\% & \multicolumn{1}{c|}{87.16\%} & 86.62\% & 88.92\% & 87.77\% \\

\multicolumn{1}{c|}{\tool{}} & 98.89\% & 97.20\% & \multicolumn{1}{c|}{98.04\%} & 92.12\% & 82.08\% & \multicolumn{1}{c|}{87.10\%} & 84.65\% & 87.60\% & \multicolumn{1}{c|}{86.13\%} & 85.77\% & 86.98\% & \multicolumn{1}{c|}{86.37\%} & 97.12\% & 82.52\% & \multicolumn{1}{c|}{89.82\%} & 91.77\% & 90.54\% & 91.15\% \\

\multicolumn{1}{c|}{\tool{}+Replay 10\%} & 98.94\% & 97.17\% & \multicolumn{1}{c|}{\textbf{98.05\%}} & 96.96\% & 81.90\% & \multicolumn{1}{c|}{\textbf{89.43\%}} & 87.22\% & 86.66\% & \multicolumn{1}{c|}{\textbf{86.94\%}} & 91.38\% & 86.02\% & \multicolumn{1}{c|}{\textbf{88.70\%}} & 98.52\% & 81.46\% & \multicolumn{1}{c|}{\textbf{89.99\%}} & 92.54\% & 90.38\% & \textbf{91.46\%} \\ \hline
\end{tabular}
}
\end{table*}

\begin{table*}[]
\centering
\caption{\zsd{Results of Model Incremental Development(Scenario II)}}
\vspace{-10pt}
\label{rq4-1}
\resizebox{\textwidth}{!}{%
\begin{tabular}{cccccccc}
\hline
 &  & Retrain & Replay 5\% & Replay 10\% & Global & Category-wise & \tool{} \\ \hline
\multirow{4}{*}{MLP-MNIST} & TaskA & 0.00\% & 91.93\% & 94.85\% & 94.22\% & 93.55\% & 98.95\% \\
 & TaskB & 0.14\% & 85.89\% & 90.71\% & 97.54\% & 98.79\% & 97.43\% \\
 & TaskC & 98.11\% & 97.96\% & 97.88\% & 97.88\% & 97.98\% & 97.93\% \\\cline{2-8} 
 & Avg & 32.75\% & 91.93\% & 94.48\% & 96.55\% & 96.77\% & \textbf{98.11\%} \\\hline
\multirow{4}{*}{MLP-FMNIST} & TaskA & 0.00\% & 58.93\% & 66.40\% & 84.90\% & 90.13\% & 93.57\% \\
 & TaskB & 0.00\% & 55.47\% & 68.63\% & 91.03\% & 96.10\% & 96.73\% \\
 & TaskC & 96.80\% & 95.58\% & 94.63\% & 96.15\% & 96.25\% & 96.15\% \\\cline{2-8} 
 & Avg & 32.27\% & 69.99\% & 76.55\% & 90.69\% & 94.16\% & \textbf{95.48\%} \\\hline
\multirow{4}{*}{AlexNet-CIFAR10} & TaskA & 0.00\% & 37.57\% & 57.17\% & 75.57\% & 87.83\% & 93.37\% \\
 & TaskB & 0.00\% & 29.30\% & 50.83\% & 59.07\% & 70.17\% & 76.03\% \\
 & TaskC & 94.63\% & 94.60\% & 93.63\% & 92.70\% & 94.10\% & 92.93\% \\\cline{2-8} 
 & Avg & 31.54\% & 53.82\% & 67.21\% & 75.78\% & 84.03\% & \textbf{87.44\%} \\\hline
\multirow{4}{*}{VGG16-CIFAR10} & TaskA & 0.00\% & 38.63\% & 54.77\% & 86.90\% & 76.57\% & 89.17\% \\
 & TaskB & 0.00\% & 33.10\% & 45.67\% & 54.33\% & 60.80\% & 65.93\% \\
 & TaskC & 94.53\% & 92.30\% & 91.30\% & 91.40\% & 89.65\% & 91.55\% \\\cline{2-8} 
 & Avg & 31.51\% & 54.68\% & 63.91\% & 77.54\% & 75.67\% & \textbf{82.22\%} \\\hline
\multirow{4}{*}{ResNet101-CIFAR10} & TaskA & 10.03\% & 17.90\% & 38.37\% & 91.90\% & 92.20\% & 92.30\% \\
 & TaskB & 6.37\% & 15.70\% & 27.63\% & 64.60\% & 57.07\% & 62.63\% \\
 & TaskC & 89.73\% & 90.70\% & 87.13\% & 86.80\% & 89.73\% & 89.35\% \\\cline{2-8} 
 & Avg & 35.38\% & 41.43\% & 51.04\% & 81.10\% & 79.66\% & \textbf{81.43\%} \\\hline
\multirow{4}{*}{VGG19-\newline ImageNet} & TaskA & 0.00\% & 31.28\% & 32.44\% & 93.33\% & 92.31\% & 98.21\% \\
 & TaskB & 0.00\% & 28.08\% & 29.74\% & 80.13\% & 83.72\% & 85.13\% \\
 & TaskC & 92.31\% & 92.02\% & 91.44\% & 93.08\% & 93.46\% & 92.60\% \\\cline{2-8} 
 & Avg & 30.77\% & 50.46\% & 51.21\% & 88.85\% & 89.83\% & \textbf{91.98\%} \\\hline
\end{tabular}
}
\end{table*}

\noindent\textbf{Detailed Configuration.} \zsd{To answer RQ4, we set up two experimental scenarios. In the first scenario, we split the datasets MNIST, FMNIST, CIFAR10, and ImageNet into two subtasks, each containing five categories, labeled as task \( A \) (previous task) and task \( B \) (new task). For example, in the CIFAR10 dataset, which consists of ten distinct image categories, task A includes the first five categories, while task B includes the remaining five categories. In the second scenario, we further divide the datasets into three subtasks. For instance, task \( A \) and task \( B \) include three categories each, while task \( C \) includes four categories.}

To simulate the process of incremental model development, for the first scenario, we start by training models exclusively on task \( A \) in six different experimental setups following the division. Then, we proceed to train these models on task \( B \) while aiming to preserve their performance on task \( A \). The evaluation focuses on the accuracy for both task \( A \) and task \( B \). Here, the accuracy on task \( A \) evaluates how well the models retain knowledge of the original task without forgetting, while the accuracy on task \( B \) gauges the models' ability to learn new tasks. \zsd{The second scenario follows a similar process (with an additional round of new task learning). Here, the model first learns task A, followed by task B, and finally task C.}

We would like to note that \emph{this setup aims to simulate the actual scenario of model incremental development, where we typically have limited access to the training data of previous tasks}. Therefore, we introduce `replay' as a baseline for comparison, which refers to relaxing the restrictions on accessing the data of previous tasks while learning a new task. This is done by revisiting some of the task \( A \) data to aid the model in remembering and maintaining its performance on the old task.

In particular, we aim to compare the following methods: naive retraining (\ie, directly training with new task data only), Replay 5\% (i.e., training data includes 5\% of previous task data along with all new task data), Replay 10\%, Replay 100\%, DeepArc (i.e., training the new task only on the redundant layers of the model), global slicing, category-wise slicing, \tool{}, and \tool{} combined with Replay. 

\noindent\textbf{Results and Findings.} The results of Experiment Scenario 1, as shown in Table \ref{rq4}, indicate that direct retraining causes catastrophic forgetting in all settings, with the model completely forgetting knowledge related to task \( A \) in six settings. Replay methods outperform direct retraining, with 5\% data replay resulting in an average 30.45\% decrease in predictive accuracy for task \( A \), 10\% data replay leading to a 23.03\% average decrease, and 100\% data replay causing a 5.15\% average decrease. \zsd{Additionally, global slicing and category-wise slicing also achieve promising results, with global slicing reducing task \( A \)'s accuracy by 11.75\% and category-wise slicing by 10.58\%, both outperforming 5\% and 10\% data replay.}

Our method outperforms 5\% and 10\% data replay, as well as global slicing and category-wise slicing, and achieves comparable performance to 100\% data replay in most scenarios, with an average precision loss of only 2.79\% on task \( A \). When combined with 10\% data replay, our method consistently achieves the best results in all experimental settings, with an average precision loss for task \( A \) of only 0.25\% and an average accuracy of 90.76\%. 

These results thoroughly demonstrate that our method effectively mitigates the issue of catastrophic forgetting in task \( A \). Even with a fixed portion of the model's parameters, our method incurs only around a 1.64\% loss in accuracy on task \( B \) compared to direct retraining. Notably, in terms of storage efficiency, replay methods require storing some training data from task \( A \), typically needing several to tens of megabytes of storage space under our experimental settings. In contrast, our method only needs to retain index information of critical neurons in the model, significantly reducing storage requirements to just a few kilobytes, which is particularly crucial for applications with limited storage resources. Moreover, in terms of efficiency, our method is slightly superior to the 5\% and 10\% data replay and is twice as efficient as the 100\% data replay, with more details provided in~\cite{NeuSemSlice2024}. Overall, considering the combined performance on tasks \( A \) and \( B \), our method demonstrates significant effectiveness in the model incremental development task.

Regarding DeepArc, we find that training on the identified redundant layers of the model does not effectively alleviate catastrophic forgetting. It maintains only partial capability to complete task \( A \) on VGG16 and ResNet101, and DeepArc leads to a substantial reduction in the model's ability to learn new tasks.

\zsd{The results for the second scenario appear in Table \ref{rq4-1}. We further compare \tool{} with naive retraining, replay-based methods, global slicing, and category-wise slicing~\footnote{Due to the significant decline in DeepArc's ability to learn new tasks observed in Experiment Scenario I, it is unable to proceed with training task C. Therefore, we do not include it in the comparison here.}. In six experimental settings, direct retraining once again forgets knowledge related to task $A$ and task $B$. Replay-based methods also decline compared to the first scenario. For example, replay that uses 5\% of the data yields average accuracies of 91.93\%, 69.99\%, 53.82\%, 54.68\%, 41.43\%, and 50.46\% across the six settings, while replay that uses 10\% of the data produces similar results and performs well only on MLP-MNIST. In contrast, \tool{} achieves the highest performance, with an overall average accuracy of 89.44\% in the six settings, which is 4.36\% higher than global slicing and 2.75\% higher than category-wise slicing. However, the introduction of more complex tasks leads to imbalanced performance across tasks, with forgetting in task B being the most severe in most settings. A possible explanation is that task A, as the base model, is deeply embedded in the model's parameters, while task C receives focused training as the newest target, leading to stronger performance on these two tasks. 

To further validate the effectiveness of \tool{}, we perform paired \( t \)-tests across all experimental settings to compare its performance with baseline methods, including replay-based methods, global slicing, and category-wise slicing. The results show that \tool{}'s improvements are statistically significant (\( p \)-values < 0.05, Cohen’s \( d \) > 0.8) in all comparisons except for replay 10\% (\( p \)-values = 0.06, Cohen’s \( d \) = 0.98) and replay 100\% (\( p \)-values = 0.33, Cohen’s \( d \) = 0.44) in Experiment Scenario 1. These statistical findings confirm the robustness and generalizability of \tool{} in incremental development scenarios, providing strong evidence of its practical utility in mitigating catastrophic forgetting while maintaining computational and storage efficiency.}

\begin{tcolorbox}[size=title]
{\textbf{Answer to RQ4: In Experiment Scenario 1, our method reduces the average accuracy loss for the original task to 2.79\% and achieves 89.77\% average accuracy in the incremental model development task. Combined with replay, it consistently excels across all settings, reducing the average accuracy loss for the original task to 0.25\% and increasing average accuracy to 90.76\%, effectively countering catastrophic forgetting. Similar trends are also observed in Experiment Scenario 2.}}
\end{tcolorbox}

\begin{figure*}[h]
    \centering
    \includegraphics[width=0.95\textwidth]{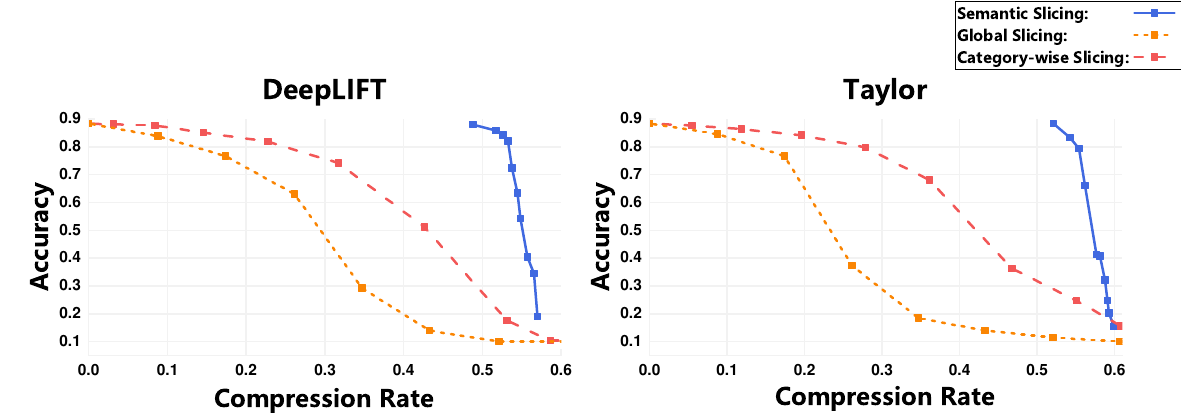}
    \caption{Comparison Results of Performance for Semantic Slicing, Global Slicing, and Category-Wise Slicing on MobileNet (Left: Based on DeepLIFT, Right: Based on Taylor). Ideally, Dots (Lines) Located at the \textbf{Upper Right} Are Favored for Their Higher Compression Rates and Better Accuracy Performance, Which Also Suggest That the Corresponding Slicing Methods Better Identify Critical Neurons in the Model.}
    \label{res:mobilenet}
\end{figure*}
\subsection{Applications in Software Engineering: A Case Study on MobileNet}
\label{sec:application}
To further demonstrate the applicability of our approach in software engineering, particularly in resource-constrained or embedded deployment scenarios, we extend our evaluation to the widely adopted MobileNet architecture~\cite{howard2017mobilenets}. MobileNet is extensively utilized in real-world environments, ranging from embedded systems to intelligent IoT devices, making it a strong representative for mission-critical deployments. Specifically, we continue using CIFAR-10 as the target dataset and adopt the latest MobileNet V3 as the target model, conducting a case study under the same experimental settings. Through this evaluation, we aim to assess whether \tool{} can consistently facilitate effective model maintenance in architectures that closely align with real-world software systems.

\noindent\textbf{Experimental Results.} Figure~\ref{res:mobilenet} presents the effectiveness comparison of semantic slicing, global slicing, and category-wise slicing on MobileNet using DeepLIFT and Taylor. The results clearly indicate that, at comparable compression rates (i.e., when identifying a similar proportion of critical neurons), semantic slicing more accurately identifies critical neurons in the model due to its semantic-based approach. This leads to improved model accuracy under equivalent compression rates. Specifically, for Taylor, when the model compression rate is approximately 50\%, semantic slicing achieves an accuracy of 88.41\%, whereas global slicing and category-wise slicing yield accuracies of 13.90\% and 36.25\%, respectively. DeepLIFT exhibits a similar trend. Additionally, global slicing and category-wise slicing maintain reasonable accuracy only when the compression rate is reduced to 20\% or lower.

Furthermore, the three slicing methods identified using Taylor are applied in Model Restructure and integrated with existing pruning techniques to evaluate their impact on model compression efficiency. Experimental results demonstrate that for Global and ERK, Model Restructure with semantic slicing enables the model to achieve a compression rate of approximately 60\% within only two pruning iterations, while maintaining an accuracy of 88.41\%. In contrast, directly applying Global or ERK requires approximately nine iterations to reach a similar compression rate and model accuracy. 
For the Uniform+ method, without semantic slicing, the model undergoes nine pruning iterations to achieve a compression rate of 61.26\%, at which point accuracy drops to 19.34\%. However, when assisted by semantic slicing in Model Restructure, only two pruning iterations are needed to reach a compression rate of 61.71\%, while accuracy remains at 87.95\%. In contrast, global slicing and category-wise slicing exhibit inaccuracies in identifying critical neurons, leading to substantial accuracy degradation during pruning~(falling below 30\%), ultimately compromising model usability.

\noindent\textbf{Conclusion and Outlook.} Our case study on MobileNet confirms that the proposed approach can be effectively adapted to resource-constrained or embedded software systems, enabling streamlined model maintenance without compromising performance. These promising results suggest broader applicability in other software engineering scenarios. In future work, we plan to extend our methodology to more diverse real-world applications to further demonstrate its utility.

\subsection{Threats to Validity}
\label{sec:threats-validity}
\subsubsection{External Threats} The threat to external validity arises from the particular settings we have chosen for our study, which raises concerns about the generalizability of our proposed approach. To mitigate these threats, we have selected diverse evaluation settings. In more detail, we select four widely utilized datasets-MNIST, Fashion-MNIST, CIFAR10, and ImageNet, and employ six classical DNN architectures-MLP, AlexNet, MobileNet,VGG16, ResNet101, and VGG19. 
The experiments are conducted under seven settings.
This selection is intended to enhance the generalizability of our findings across common scenarios within the domain.

\subsubsection{Internal Threats} Internal validity threats stem from the tools utilized in our research, which include captum~\cite{kokhlikyan2020captum} and cleverhans~\cite{papernot2018cleverhans}. Additionally, there is a potential threat concerning the accurate reproduction of the contribution assessment metrics and the model modularization framework DeepArc in PyTorch~\cite{pytorch}. The use of various deep learning frameworks, such as TensorFlow~\cite{tensorflow}, may impact the results. Additionally, the inherent randomness in model training poses a potential threat to the internal validity. We addressed this issue by repeating key experiments more than five times and reporting the mean values. 
Finally, as mentioned in Section~\ref{sec:re-adaptation}, the resilience of the model's re-adaptation to adversarial attacks requires enhancement. This issue, though orthogonal to the main focus of this paper, highlights the need for future work to tackle and reduce its impact.

%% file: Chapters/conclusion.tex
\section{Conclusion}
This paper introduces \tool{}, an innovative framework that integrates semantic slicing with semantic-aware model maintenance for deep neural networks. Semantic slicing identifies critical neurons at a granular level, categorizing and merging them across layers based on semantic similarities. This approach markedly enhances the effectiveness of these neurons in model maintenance tasks, surpassing global slicing and category-wise slicing. \tool{} further innovates in semantic-aware model maintenance by offering strategies for model restructure that preserve critical neurons, efficient re-adaptation through tuning only these neurons, and incremental development that maintains critical neurons while training non-critical ones. The effectiveness of \tool{} is demonstrated through comprehensive evaluations, showing its superiority over baselines in all three aspects of model maintenance.

\section*{Acknowledgments}
We thank the editor and anonymous reviewers for their constructive comments. This work was supported by the National NSF of China (Grant Nos. 62302176, 62072046, 62302181), the Key R\&D Program of Hubei Province (2023BAB017, 2023BAB079), and the Knowledge Innovation Program of Wuhan-Basic Research (2022010801010083).